\documentclass{article}

\usepackage{arxiv}
\usepackage[utf8]{inputenc}
\usepackage[T1]{fontenc}
\usepackage{amsfonts}
\usepackage{microtype}
\usepackage{nicefrac}

\usepackage{amsmath}
\usepackage{amsthm}
\usepackage{amsbsy}
\usepackage{tikz}
\usepackage{booktabs}
\usepackage{tabularx}
\usepackage{array}
\usepackage{multirow}
\usepackage{chngcntr}

\usepackage{graphicx}
\usepackage{float}
\usepackage{subcaption}
\usepackage{wrapfig}

\usepackage{xcolor}
\usepackage{pifont}

\newcommand{\cmark}{\ding{51}}
\newcommand{\xmark}{\ding{55}}

\usepackage{algorithm}
\usepackage{algpseudocode}

\usepackage[numbers,square]{natbib}
\usepackage{url}
\usepackage{hyperref}

\newtheorem{definition}{Definition}[section]
\newtheorem{theorem}{Theorem}[section]

\newtheorem{example}{Example}[section]
\newtheorem{remark}{Remark}[section]
\newtheorem{proposition}{Proposition}

\title{Beyond the Lower-Star Filtration: Alternative Filtration Methods for Persistent Homology}

\author{
    Innocent Abaa\\
    Department of Mathematics\\
    Juniata College\\
    \texttt{iga100@juniata.edu} \\
        \And
   Umar Islambekov\\
    Department of Mathematics and Statistics\\
    Bowling Green State University\\
    \texttt{iumar@bgsu.edu} \\
}

\begin{document}
\maketitle

\vspace{-0.5cm}

\begin{abstract} In the lower-star filtration (LSF), simplices are assigned filtration values according to the maximum-vertex rule, with each simplex entering the filtration when its highest-valued vertex appears. A limitation of this construction is that some higher-dimensional simplices may enter simultaneously with their boundary faces, causing certain topological features to be filled immediately upon formation and thus not appear in the resulting persistence diagram. To address this limitation, we propose three alternative filtration methods: the \(\varepsilon\)-lower-star filtration (ELSF), the \(L_p\) filtration (LpF), and the softmax filtration (SMF). {ELSF introduces a dimension-dependent delay to the maximum-vertex rule,} allowing higher-dimensional simplices to enter the filtration after their boundary faces, whereas {LpF and SMF use aggregation functions that depend on all vertex values rather} than solely on their maximum. By modifying how filtration values are assigned, these methods allow simplices to enter the filtration more progressively, so that topological features such as loops and voids may persist over longer filtration intervals rather than being filled immediately. Consequently, the resulting persistence diagrams may include persistence features that are not observed under LSF, potentially providing useful information for distinguishing data populations. Experimental results on simulated and real-world datasets demonstrate that the proposed filtration methods are computationally efficient and can provide greater sensitivity to structural differences in the data than LSF.
\end{abstract}


\section{Introduction}
\label{sec1}
In recent years, graph data have become one of the most actively studied data types in modern data analysis. This growing interest is largely driven by the ability of graphs to naturally represent complex relational systems across diverse domains, including social networks, biological systems, chemistry, communication networks, and~information science~\citep{Nettleton2013, Sun2017}. In~these settings, entities are represented as nodes (vertices), while relationships or interactions between entities are represented as edges. Such representations provide a flexible and expressive framework for modeling dependency structures that are often difficult to capture using traditional vector-based data formats. As~increasingly large and complex datasets continue to emerge, the~analysis of graph data has become an important area of research, creating a growing need for methodologies capable of extracting meaningful structural information from these systems in both mathematically principled and computationally efficient~ways.

Topological Data Analysis (TDA) provides a promising framework for studying such complex data by offering mathematical tools for quantifying the underlying shape and structural organization of data~\cite{carlsson2009topology}. At~the heart of TDA lies persistent homology (PH), a~central tool that tracks the evolution of topological features such as connected components, loops, and~higher-dimensional voids across multiple scales \citep{ZomorodianCarlsson2005}. This is achieved through the construction of a filtration, a~nested sequence of abstract simplicial complexes built on top of data. From~a practical perspective, simplicial complexes (geometric objects built from vertices, edges, triangles, and~their higher-dimensional generalizations) serve as a bridge between discrete observations and the underlying continuous structure from which the data are sampled \citep{nanda2014simplicial}. While PH has been highly successful for data types such as point clouds and images, where a natural notion of distance exists between observations~\cite{Ghrist2008}, extending these ideas to graph data is not straightforward because graphs generally lack a natural intrinsic metric space~\cite{aktas2019persistence}. 

One prominent approach for extending PH to graph data is to first embed the graph into a metric space and treat the embedded graph as a geometric point cloud, thereby allowing distance-based filtrations such as the Vietoris–Rips or Čech filtrations to be applied. A~representative example is the framework proposed by You~et~al.~\cite{you2021topics}, which employs a latent space model~\cite{hoff2002latent} to embed graphs into a Euclidean space prior to applying persistent homology. Although~this approach is mathematically appealing, it introduces several important limitations. First, the~embedding process may distort the intrinsic topology of the original graph. Graphs are fundamentally relational structures whose properties are determined by connectivity patterns among vertices and edges; however, embedding replaces these relational relationships with geometric approximations, potentially obscuring important topological features. Second, the~resulting topological representation depends heavily on modeling assumptions, including the choice of latent space, embedding dimension, distance metric, and~parameter estimation procedures, such as maximum likelihood estimation or Markov chain Monte Carlo methods. Consequently, different modeling choices may produce substantially different embeddings and, therefore, different topological summaries. The~overall framework can also become computationally expensive for large-scale graphs. In~particular, latent space embedding methods are often costly to estimate, while the construction of Vietoris--Rips complexes becomes increasingly expensive due to the rapid growth in the number of simplices as the number of vertices~increases.

An alternative approach is to analyze graphs directly using quantitative node attributes defined on the vertices of the graph. In~this setting, topological information is extracted through the lower-star filtration (LSF) \cite{edelsbrunner2022computational}, which constructs filtrations directly on the graph without requiring an intermediate geometric embedding~\cite{AbaaUmar2025,islambekov2023fasttopologicalapproachpredicting}. This approach preserves the intrinsic relational structure of the graph while remaining computationally efficient. However, the~LSF suffers from an important limitation in the way simplices enter the filtration. Under~the lower-star filtration, a~simplex \(\sigma\) enters according to the maximum function value among its vertices, \(f(\sigma)=\max_{v\in\sigma} f(v)\), causing simplices to appear at the same filtration level as their highest-valued vertex. As~a result, higher-dimensional simplices, such as triangles, may enter at the same filtration value as their boundary edges. For~example, a~loop formed by three connected edges can be filled immediately when the corresponding \(2\)-simplex enters at the same filtration value. Consequently, the~resulting persistence diagrams, which summarize the birth and death of topological features across the filtration, may contain only a limited number of higher-dimensional features, as~some features may be filled immediately upon formation and thus not appear in the resulting persistence~diagrams.

In this paper, we propose three alternative filtration methods: the \(\varepsilon\)-lower-star filtration (ELSF), the~\(L_p\) filtration (LpF), and~the softmax filtration (SMF). {These methods generalize the lower-star filtration by introducing alternative rules for assigning filtration values to simplices. While ELSF modifies the maximum-vertex
rule through a dimension-dependent adjustment, LpF and SMF incorporate the values of all vertices when determining simplex filtration values. These modifications may allow higher-dimensional simplices to enter the filtration after their boundary faces, enabling certain topological features, such as loops and voids, to~persist over longer filtration
intervals. Consequently, the~resulting persistence diagrams may contain additional persistence features that are not observed under the lower-star filtration. These additional features may provide useful information for distinguishing structural differences between data populations, potentially improving the discriminative power of persistence diagrams while preserving the computational advantages of direct graph-based~analysis.

In statistical machine learning, hypothesis testing on graph data aims to determine whether distinct graph populations arise from the same underlying mechanism or from different random graph models. A~common framework in TDA represents each graph through its persistence diagram and compares graph populations using pairwise distances between these diagrams. Since persistence diagrams do not possess natural parametric distributions, Robinson and Turner}~\cite{robinson2017hypothesis} {proposed a nonparametric permutation-based hypothesis testing framework that uses a joint loss function based on within-group bottleneck or Wasserstein distances between persistence diagrams as the test statistic. However, this approach has two characteristics that motivate our modifications. Although~efficient approximation algorithms have been developed for computing bottleneck and Wasserstein distances, these computations still require solving optimal matching problems and can remain computationally demanding, particularly for large datasets. As~a result, the~overall permutation framework may be difficult to scale. In~addition, the~underlying loss function primarily measures within-group dispersion without explicitly accounting for between-group separation.}

In this work, we propose two modifications to the framework of~\cite{robinson2017hypothesis}. First, we replace the original joint loss function, which is based solely on within-group pairwise distances between persistence diagrams, with~the energy statistic (\(E\)-statistic) \cite{szekely2013energy}. {This allows the testing procedure to incorporate both within-group variability and between-group differences, providing an alternative formulation for detecting distributional discrepancies between graph populations. Second, rather than comparing persistence diagrams directly through bottleneck or Wasserstein distances, we follow the approach of} \cite{islambekov2024vector} {and compare their vectorized summaries. This avoids the computational burden associated with bottleneck and Wasserstein distance calculations, thereby improving computational efficiency while providing a framework for hypothesis testing based on vectorized persistence diagram summaries.}

The remainder of this paper is organized as follows. Section~\ref{sec2} reviews the background of TDA and its fundamental tools. Section~\ref{sec3} presents the proposed filtration methods. Section~\ref{sec4} reviews the hypothesis testing procedure for comparing persistence diagrams developed by~\cite{robinson2017hypothesis} and presents the proposed modifications. Section~\ref{sec5} reports the results of numerical experiments comparing the proposed filtration methods with the  lower-star filtration and latent space embedding approaches. Finally, Section~\ref{sec13} concludes with a discussion of the main findings and directions for future~research.

\section{Background and~Preliminaries}\label{sec2}
\unskip

\subsection{Persistent~Homology}
The theoretical foundation of persistent homology lies in algebraic topology, a~branch of mathematics that uses algebraic tools to study topological spaces~\cite{hatcher2002algebraic}. One of the main goals of algebraic topology is to study topological invariants-features such as connected components, loops, and~higher-dimensional voids-that remain unchanged under continuous deformations such as stretching, compression, and~rotation. Because~these invariants are preserved under continuous transformations, they provide a rigorous way to quantify the shape structure
underlying data. TDA adapts these ideas to data analysis by developing mathematical and computational tools for studying the geometric and topological properties of complex datasets. In~particular, persistent homology extends  homology to data settings by tracking the evolution of topological features across multiple~scales.

To perform this multiscale analysis computationally, the~data are first represented by a combinatorial structure called a \emph{simplicial complex}. Simplicial complexes provide discrete approximations of topological spaces and serve as a bridge between discrete data and continuous topology. Formally, an~abstract simplicial complex \(\mathcal{K}\) on a vertex set \(\mathcal{V}\) is a collection of finite subsets of \(\mathcal{V}\) satisfying two conditions: (i) for every vertex \(v \in \mathcal{V}\), the~singleton set \(\{v\}\) belongs to \(\mathcal{K}\); (ii) if \(\tau \in \mathcal{K}\), then every subset \(\sigma \subseteq \tau\), called a face of \(\tau\), also belongs to \(\mathcal{K}\). A~simplex \(\tau \in \mathcal{K}\) containing \(k+1\) vertices is called a \(k\)-simplex and has dimension \(k\). The~dimension of a simplicial complex is the maximum dimension of its~simplices.

\begin{figure}[H]
\centering
\includegraphics[width=0.7\textwidth]{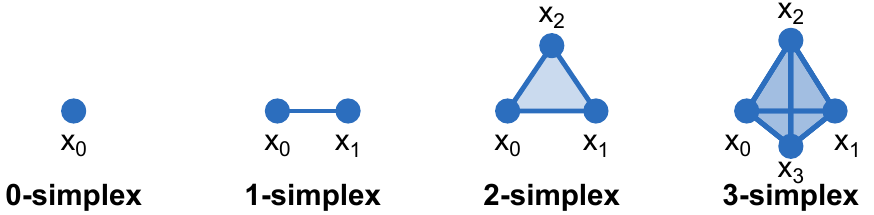}
\caption{Geometric representations of simplices. From~left to right: a $0$-simplex (point), a~$1$-simplex (edge), a~$2$-simplex (filled triangle), and~a $3$-simplex (tetrahedron).}
\label{fig:simplices}
\end{figure}

An abstract simplicial complex can be realized geometrically by assigning a geometric simplex to each abstract simplex in such a way that the face relations are preserved. In~this realization, a~$0$-simplex is represented by a point, a~$1$-simplex by a line segment, a~$2$-simplex by a triangle, and~a $3$-simplex by a tetrahedron (see Figure~\ref{fig:simplices}); higher-dimensional simplices are natural generalizations of these objects. These geometric simplices serve as the fundamental building blocks of a geometric simplicial~complex.

To study the topology of data across multiple scales, a~filtration is constructed as a nested sequence of simplicial complexes satisfying
\[
\emptyset \subseteq \mathcal{K}_1 \subseteq \mathcal{K}_2 \subseteq \cdots \subseteq \mathcal{K}_N = \mathcal{K}.
\]

As the filtration parameter increases, simplices are added sequentially to the complex, allowing topological features such as connected components, loops, and~higher-dimensional voids to appear and disappear across multiple~scales.

More generally, filtrations can be constructed through sublevel sets of a monotone function defined on a simplicial complex~\cite{nanda2014simplicial}. Let \(g:\mathcal{K}\rightarrow\mathbb{R}_{\geq 0}\) be a function satisfying the monotonicity condition $\sigma\subseteq\tau \;\Longrightarrow\; g(\sigma)\le g(\tau),$ for all simplices \(\sigma,\tau\in\mathcal{K}\). Let \(c_1<c_2<\cdots<c_N\) denote the distinct function values attained by \(g\). Then \(g\) induces the~filtration \[\emptyset\subseteq\mathcal{K}_{c_1}\subseteq\mathcal{K}_{c_2}\subseteq\cdots\subseteq\mathcal{K}_{c_N}=\mathcal{K},\] where $\mathcal{K}_{c_i}=\{\sigma\in\mathcal{K}\mid g(\sigma)\le c_i\}, \qquad i=1,\ldots,N.$

Different choices of the monotone function $g$ produce different filtrations depending on the type of data and the structural properties one aims to capture. For~point cloud data equipped with a metric, two commonly used constructions are the Vietoris--Rips and \v{C}ech filtrations. Let $P=\{p_1,p_2,\dots,p_n\}\subseteq\mathbb{R}^d$ be a finite point cloud with metric $d(\cdot,\cdot)$. The~Vietoris--Rips complex at scale $\epsilon\ge0$, denoted by $\mathrm{VR}_{\epsilon}(P)$, is defined as
$\mathrm{VR}_{\epsilon}(P)=\{\sigma\subseteq P \mid d(p,q)\le\epsilon \text{ for all } p,q\in\sigma\}.$ Thus, a~simplex belongs to the Vietoris--Rips complex whenever all pairwise distances between its vertices are at most $\epsilon$. Defining
$g_V(\sigma)=\max_{p,q\in\sigma} d(p,q),$
the monotonicity condition 
induces the Vietoris--Rips filtration
$\mathrm{VR}_{\epsilon_1}(P)\subseteq\mathrm{VR}_{\epsilon_2}(P)\subseteq\cdots\subseteq\mathrm{VR}_{\epsilon_N}(P),$ where $\epsilon_1<\epsilon_2<\cdots<\epsilon_N$ are the distinct pairwise distances between points in $P$.

Similarly, the~\v{C}ech complex at scale $\epsilon\ge0$, denoted by $\check{C}_{\epsilon}(P)$, is defined as
$\check{C}_{\epsilon}(P)=\left\{\sigma\subseteq P \mid \bigcap_{p\in\sigma} B_{\epsilon}(p)\neq\emptyset\right\}.$
Hence, a~simplex is included whenever the corresponding $\epsilon$-balls centered at its vertices possess a nonempty common intersection. Defining $g_C(\sigma)=\min\left\{r\ge0 \mid \exists x\in\mathbb{R}^d \text{ such that } d(x,p)\le r \text{ for all } p\in\sigma\right\},$
the monotonicity condition 
induces the \v{C}ech filtration
$\check{C}_{\epsilon_1}(P)\subseteq\check{C}_{\epsilon_2}(P)\subseteq\cdots\subseteq\check{C}_{\epsilon_N}(P).$ As the scale parameter increases, both the Vietoris--Rips and \v{C}ech filtrations generate nested sequences of simplicial complexes that allow persistent homology to study the evolution of topological structure across multiple~scales.

For graph and image data, filtrations can also be constructed using real-valued functions defined on the vertices rather than pairwise distances. One commonly used construction is the lower-star filtration (LSF). Let \(X\) denote the vertex set of a simplicial complex \(\mathcal{K}\), and~let \(f:X\rightarrow\mathbb{R}_{\geq 0}\) be a real-valued function defined on the vertices. The~lower-star filtration extends \(f\) to simplices by assigning to each simplex \(\sigma\in\mathcal{K}\) the filtration value
$f(\sigma)=\max_{v\in\sigma} f(v).$
Thus, a~simplex enters the filtration when its highest-valued vertex appears. Since
$\sigma\subseteq\tau
\Longrightarrow
f(\sigma)\le f(\tau),$
the function \(f\) induces the filtration $\emptyset
\subseteq
\mathcal{K}_{c_1}
\subseteq
\mathcal{K}_{c_2}
\subseteq
\cdots
\subseteq
\mathcal{K}_{c_N}
=
\mathcal{K},$
where
$
\mathcal{K}_{c_i}
=
\{\sigma\in\mathcal{K}\mid f(\sigma)\le c_i\},$
and \(c_1<c_2<\cdots<c_N\) are the distinct function values attained by \(f\). Unlike distance-based filtrations such as the Vietoris--Rips and \v{C}ech filtrations, the~lower-star filtration does not require the data to lie in a metric space.  Instead, the~filtration is determined entirely by the vertex-based function \(f\), making the construction particularly useful for graph and image data where meaningful pairwise distances may not be naturally~defined.

Once a filtration has been constructed, homology provides the algebraic framework for quantifying the topological features that emerge throughout the filtration. Homology associates to a simplicial complex \(\mathcal{K}\) a sequence of algebraic objects called homology groups, denoted by \(H_d(\mathcal{K})\), where \(d\) represents the topological dimension. These groups are constructed from chain groups and boundary operators that encode relationships between simplices of adjacent dimensions. For~each dimension \(d\), the~\(d\)-dimensional chain group \(C_d(\mathcal{K})\) consists of formal linear combinations of \(d\)-simplices, while the boundary operator $\partial_d:C_d(\mathcal{K})\rightarrow C_{d-1}(\mathcal{K})$ maps each simplex to the formal sum of its \((d-1)\)-dimensional faces. A~fundamental property of the boundary operator is that $\partial_{d-1}\circ\partial_d=0,$ meaning that the boundary of a boundary is always empty. This property allows the definition of the cycle group $Z_d(\mathcal{K})=\ker(\partial_d),$
consisting of \(d\)-chains with no boundary, and~the boundary group $B_d(\mathcal{K})=\mathrm{im}(\partial_{d+1}),$
consisting of \(d\)-chains that arise as boundaries of higher-dimensional simplices. Since every boundary is also a cycle, that is, $B_d(\mathcal{K})\subseteq Z_d(\mathcal{K}),$
the \(d\)-th homology group is defined as the quotient space
\[
H_d(\mathcal{K})=Z_d(\mathcal{K})/B_d(\mathcal{K}).
\]

Intuitively, homology identifies cycles that are not themselves boundaries of higher-dimensional simplices, thereby capturing topological holes within the simplicial complex. The~dimension of \(H_d(\mathcal{K})\), called the \(d\)-th Betti number and denoted by \(\beta_d\), quantifies the number of independent topological features in dimension \(d\). In~particular, \(\beta_0\) counts connected components, \(\beta_1\) counts loops or tunnels, and~\(\beta_2\) counts higher-dimensional voids or~cavities.

Persistent homology extends this construction to filtrations by computing homology across a nested sequence of simplicial complexes $\emptyset=\mathcal{K}_0\subseteq\mathcal{K}_1\subseteq\cdots\subseteq\mathcal{K}_n=\mathcal{K}.$ The inclusion maps between simplicial complexes induce homomorphisms between their corresponding homology groups,
\(
\iota_d^{a\rightarrow b}:H_d(\mathcal{K}_a)\rightarrow H_d(\mathcal{K}_b), a\le b,
\)
allowing topological features to be tracked throughout the filtration. As~simplices are progressively added, topological features appear and later disappear through merging or being filled in. Each feature is therefore associated with a birth time \(b_i\), representing the filtration scale at which it first appears, and~a death time \(d_i\), representing the scale at which it disappears. The~persistence of a feature is measured by the interval length $d_i-b_i,$ where long intervals typically correspond to meaningful structural information, while short intervals are often interpreted as topological noise~\cite{edelsbrunner2022computational}.

The collection of birth--death intervals is commonly summarized through two standard representations: persistence barcodes and persistence diagrams. A~persistence barcode represents each topological feature as an interval \([b_i,d_i)\), while a persistence diagram represents the same information as a multiset of points \((b_i,d_i)\) in $\mathbb{R}^2$. These representations provide stable multiscale summaries of the underlying topological structure of the data and form the primary objects used in statistical and machine learning applications of TDA~\cite{chazal2021introduction}.

\subsection{Distance Metrics and Stability of Persistence~Diagrams}
Persistence diagrams provide compact multiscale summaries of the underlying topological structure of data. In~many statistical and machine learning applications, however, the~objective extends beyond simply extracting topological features. One often seeks to compare datasets, quantify structural similarity between samples, perform clustering or classification, or~conduct statistical inference based on their topological signatures. Consequently, it is necessary to equip the space of persistence diagrams with suitable notions of distance that allow these structures to be compared~quantitatively.

Since persistence diagrams are multisets of points in $\mathbb{R}^2$ rather than elements of a Euclidean vector space, specialized metrics are required to compare them. A~common approach is to define distances through optimal matchings between points in two persistence diagrams, \(D_1\) and \(D_2\). One of the most widely used metrics is the \(q\)-th Wasserstein distance,
\begin{equation}
W_q(D_1,D_2)
=
\left[
\inf_{\gamma:D_1\to D_2}
\sum_{p\in D_1}
\|p-\gamma(p)\|_\infty^q
\right]^{1/q},  
\end{equation}
where \(\gamma\) denotes a bijection between the point sets of the two diagrams. Since persistence diagrams may contain different numbers of points, unmatched points are allowed to map to their orthogonal projection on the diagonal \(d=b\), corresponding to topological features with zero~persistence.

A limiting case of the \(q\)-Wasserstein distance as \(q\to\infty\) is the Bottleneck distance,
\begin{equation}
d_{B}(D_{1},D_{2})
=
\inf_{\gamma:D_{1}\rightarrow D_{2}}
\sup_{p\in D_{1}}
\|p-\gamma(p)\|_{\infty},   
\end{equation}
which measures the largest discrepancy among all matched pairs of points, including points matched to the diagonal. The~Bottleneck distance plays a fundamental role in TDA due to the celebrated Stability Theorem~\cite{cohen2007stability}. Specifically, for~any two continuous tame functions, \(f\) and \(g\), defined on a triangulable space \(X\),
\begin{equation}
d_{B}(D(f),D(g))
\le
\|f-g\|_{\infty}.  
\end{equation}
Here, a~function is \emph{tame} if it has finitely many homological critical values and the homology groups of all its sublevel sets, $X_a=f^{-1}((-\infty,a]),$ are finitely generated. This condition ensures that the corresponding persistence diagrams contain only finitely many off-diagonal points, making the bottleneck distance well defined. Consequently, the~Stability Theorem guarantees that small perturbations in the input function produce only small changes in the corresponding persistence diagrams, establishing persistent homology as a robust framework for topological data~analysis.

Despite these desirable theoretical properties, exact computation of matching-based distances remains computationally expensive~\cite{eth_tda_ch6,lacombe2018large}. In~particular, computation of the Bottleneck distance scales approximately as \(O(n^{1.5}\log n)\), while computation of the Wasserstein distance often requires \(O(n^3)\) operations through variants of the Hungarian algorithm. As~persistence diagrams become increasingly large, these computational costs create significant challenges for large-scale statistical and machine learning applications. Although~approximate methods based on entropic regularization and Sinkhorn-type algorithms~\cite{kerber2017geometry} can substantially improve computational efficiency, persistence diagrams still lack the algebraic and geometric structure of a Hilbert space, limiting their direct integration into many standard machine learning frameworks~\cite{bubenik2020embeddings, carriere2019metric}.

Consequently, several approaches have been developed to transform persistence diagrams into representations better suited for statistical learning. Broadly speaking, these approaches either define kernels directly on the space of persistence diagrams or map persistence diagrams into vectorized representations~\cite{chung2019persistence}. In~this work, we focus on \emph{vectorization methods} for persistence diagrams, which transform persistence diagrams into finite-dimensional vector representations suitable for statistical analysis and machine learning. A~common approach is to first construct a summary function from a persistence diagram and then discretize the resulting function to obtain a finite-dimensional vector. These representations enable efficient computation and facilitate integration with a broad range of statistical and machine learning~methods.

\subsection{Summary Functions of Persistence~Diagrams}\label{sub}
In this section, we review several commonly used summary functions for persistence diagrams and discuss how they transform topological information extracted through persistent homology into finite-dimensional representations suitable for statistical analysis and machine learning~applications.

\begin{definition}[Betti Function]
The Betti function, also referred to as the \emph{Betti curve} \cite{chazal2021introduction}, provides a functional summary of a persistence diagram by counting the number of topological features that persist at each filtration value. Let $\mathcal{D}
=
\{(b_i,d_i)\}_{i=1}^N$
denote a persistence diagram in homological dimension \(k\), where \(b_i\) and \(d_i\) represent the birth and death times of the \(i\)-th topological feature, respectively. The~Betti function \(\beta_k(t)\) is defined as
\begin{equation}
\beta_k(t)
=
\sum_{i=1}^{N}
\mathbf{1}_{[b_i,d_i)}(t),
\qquad
\mathbf{1}_{[b_i,d_i)}(t)
=
\begin{cases}
1, & b_i \le t < d_i,\\
0, & \text{otherwise},
\end{cases}
\label{eq:betti_func}
\end{equation}
where \(\mathbf{1}_{[b_i,d_i)}(t)\) denotes the indicator function associated with the interval \([b_i,d_i)\). Intuitively, \(\beta_k(t)\) counts the number of \(k\)-dimensional topological features that are alive at filtration value \(t\).

To obtain a finite-dimensional vector representation suitable for statistical and machine learning applications, the~Betti function may be evaluated over a discrete sequence of filtration values $\{t_1,t_2,\dots,t_n\},$
yielding the pointwise vectorization
\begin{equation}
\left(
\beta_k(t_1),
\beta_k(t_2),
\dots,
\beta_k(t_n)
\right)
\in
\mathbb{R}^n.
\label{eq:pointwise}
\end{equation}

Alternatively, to~improve robustness against small perturbations and topological noise, one may construct an interval-based vectorization by averaging the Betti function over consecutive intervals~\cite{islambekov2024vector}:
\begin{equation}
\left(
\frac{1}{\Delta t_1}\int_{t_1}^{t_2}\beta_k(t)\,dt,
\frac{1}{\Delta t_2}\int_{t_2}^{t_3}\beta_k(t)\,dt,
\dots,
\frac{1}{\Delta t_{n-1}}\int_{t_{n-1}}^{t_n}\beta_k(t)\,dt
\right)
\in
\mathbb{R}^{n-1},
\label{eq:interval}
\end{equation}
where $\Delta t_i=t_{i+1}-t_i.$
Since \(\beta_k(t)\) is piecewise constant, these integrals are computationally efficient to evaluate and provide stable summaries of topological evolution across the filtration.
\end{definition}

\begin{definition}[Persistence Landscape]

Persistence landscapes~\cite{bubenik2015statistical} provide functional representations of persistence diagrams by transforming topological features into collections of real-valued functions. Let $D=\{(b_i,d_i)\}_{i=1}^N$ denote a persistence diagram in a fixed homological dimension. For~each persistence pair \((b_i,d_i)\), we define an associated tent function
\begin{equation}
\Lambda_i(t)=
\begin{cases}
t-b_i,
&
t\in\left[b_i,\dfrac{b_i+d_i}{2}\right],
\\[6pt]
d_i-t,
&
t\in\left[\dfrac{b_i+d_i}{2},d_i\right],
\\[6pt]
0,
&
\text{otherwise}.
\end{cases} 
\end{equation}

Each tent function is centered at the midpoint $\frac{b_i+d_i}{2},$ and attains maximum height $\frac{d_i-b_i}{2},$ which is proportional to the persistence of the corresponding topological~feature.

At each filtration value \(t\), the~j-{th} persistence landscape records the dominant tent functions through the collection
\[
\lambda_j(t)
=
j\text{-}\max
\{
\Lambda_i(t):i=1,\dots,N
\},
\qquad
j=1,2,\dots,
\]
where \(j\text{-}\max\) denotes the \(j\)-th largest value among all tent functions evaluated at \(t\). Thus, \(\lambda_1(t)\) captures the most prominent topological feature at scale \(t\), while higher-order landscapes encode less dominant~features.

For practical applications, each landscape function is typically evaluated over a discrete grid $t_1<t_2<\cdots<t_n,$
yielding the vector representation
\(
(\lambda_j(t_1),\lambda_j(t_2),\dots,\lambda_j(t_n))
\in
\mathbb{R}^n.
\)
   
\end{definition}  
\begin{definition}[Generalized (Kernel) Persistence Landscapes]

Generalized persistence landscapes extend the  persistence landscape by replacing piecewise linear tent functions with smooth localized kernel functions while preserving the same overall construction principle~\cite{berry2020functional}. Let $K:\mathbb{R}\rightarrow\mathbb{R}$
be a symmetric kernel function with compact support on \([-1,1]\) satisfying \(K(0)\neq0\), and~let \(h>0\) denote a bandwidth parameter. For~a persistence pair
$p=(b,d),$
define the kernel-based landscape~function
\begin{equation}
\Lambda_p^{(K)}(t)
=
\begin{cases}
\dfrac{d-b}{2K_h(0)}
K_h\!\left(
t-\dfrac{b+d}{2}
\right),
&
\left|
\dfrac{t-\dfrac{b+d}{2}}{h}
\right|
\le1,
\\[10pt]
0,
&
\text{otherwise},
\end{cases}
\end{equation}
where $K_h(u)=\frac{1}{h}K\!\left(\frac{u}{h}\right)$ denotes the rescaled kernel~function.

The \(j\)-th generalized persistence landscape is then defined as
\[
\lambda_j^{(K)}(t)
=
j\text{-}\max
\{
\Lambda_p^{(K)}(t):p\in D
\},
\qquad
j=1,2,\dots.
\]

The scaling ensures that $\Lambda_p^{(K)}
\left(
\frac{b+d}{2}
\right)
=
\frac{d-b}{2},$
so that the maximum height remains proportional to the persistence of the corresponding feature. Common choices of \(K\) include the triangle, Epanechnikov, and~tricubic kernels. By~replacing tent functions with smooth kernel functions, generalized persistence landscapes yield differentiable functional summaries that are well suited for statistical inference, optimization, and~machine learning applications.
\end{definition} 

\section{The New Filtration~Methods}\label{sec3}

In TDA, topological features are extracted by constructing filtrations on simplicial complexes. For~graph data, let $G=(V,E)$ be a finite, simple, undirected graph with vertex set \(V\) and edge set \(E\). The~graph is transformed into a simplicial complex by considering its complete subgraphs. Specifically, the~\emph{clique complex} (or \emph{flag complex}) of \(G\), denoted by \(K\), is the simplicial complex whose simplices are the cliques of \(G\). Thus, each vertex forms a \(0\)-simplex, each edge forms a \(1\)-simplex, each triangle forms a \(2\)-simplex, and, more generally, every complete subgraph on \(k+1\) vertices forms a \(k\)-simplex.

Let $f:V\rightarrow\mathbb{R}_{\ge0}$
be a nonnegative real-valued function defined on the vertices of \(G\). A~filtration is obtained by extending the vertex function \(f\) from the vertices to all simplices of the clique complex. The~extended function determines the order in which simplices enter the filtration and must satisfy the monotonicity condition $\tau\subseteq\sigma
\Longrightarrow
f(\tau)\leq f(\sigma),$ for every pair of simplices \(\tau,\sigma\in K\). This condition guarantees that every simplex enters the filtration only after all of its faces have already~appeared.

A widely used construction is the \emph{lower-star filtration}, in~which the filtration value assigned to a simplex \(\sigma\in K\) is defined by the maximum function value among its vertices,
\begin{equation}
f(\sigma)=\max_{v\in\sigma}f(v).
\end{equation}

Under this construction, a~simplex enters the filtration as soon as its highest-valued vertex appears. The~lower-star filtration is computationally efficient and has become one of the standard filtration methods in graph-based topological data~analysis.

{Despite its simplicity, the~lower-star filtration has important limitations. First, since the filtration value of a simplex depends only on its maximum vertex value, the~values of its remaining vertices do not influence its filtration value. Second, the~maximum-vertex rule may cause higher-dimensional simplices to enter the filtration simultaneously with some of their boundary faces, allowing topological features to be filled immediately upon formation. As~a result, some topological features may have zero persistence and therefore may not appear in the resulting persistence diagrams, while others may persist only over short filtration~intervals.

Motivated by these limitations, we propose three alternative filtration methods that assign filtration values to simplices using different rules. These constructions provide greater flexibility in simplex inclusion while preserving the monotonicity required for a valid filtration. Consequently, the~proposed methods may produce persistence diagrams containing additional persistence features that are not observed under LSF, allowing the evolution of topological structures to be observed more clearly throughout the filtration.}

\begin{definition}
Let \(K\) be a simplicial complex with vertex set \(V\), and~let $f:V\rightarrow\mathbb{R}_{\ge0}$
be a nonnegative real-valued function defined on the vertices. For~every simplex \(\sigma\in K\), we define the following filtration~functions.

\begin{enumerate}
\item \textbf{The \(\varepsilon\)-Lower-Star Filtration (ELSF):} This filtration modifies the lower-star filtration by introducing a dimension-dependent delay term \(\varepsilon>0\). The~additional term ensures that higher-dimensional simplices enter the filtration slightly after their lower-dimensional faces, thereby preventing the immediate filling of topological features. The~filtration is defined as
\begin{equation}
f_{\varepsilon}(\sigma)
=
\max_{v\in\sigma}f(v)
+
\varepsilon\,\dim(\sigma).
\end{equation}

\item \textbf{The \(\boldsymbol{L_p}\) Filtration (LpF):}  
This filtration replaces the maximum operator in the lower-star filtration with an \(L_p\)-based aggregation of the vertex values. Rather than depending only on the largest vertex value, the~filtration incorporates information from all vertices within the simplex. The~filtration is defined as
\begin{equation}
f_p(\sigma)
=
\left(
\sum_{v\in\sigma} f(v)^p
\right)^{1/p},
\qquad p\ge1.
\end{equation}

 \item \textbf{The Softmax Filtration (SMF):}  
This filtration replaces the maximum operator with a smooth softmax-based aggregation function. The~construction incorporates information from all vertices within the simplex while assigning larger weights to higher vertex values. The~filtration is defined as
\begin{equation}
f_{\beta}(\sigma)
=
\frac{1}{\beta}
\log
\left(
\sum_{v\in\sigma}
e^{\beta f(v)}
\right),
\qquad \beta>0.
    \end{equation}
\end{enumerate}
\end{definition}

\begin{remark}
The proposed filtration methods are closely related to the lower-star filtration. In~particular, the~$\varepsilon$-Lower-Star filtration reduces exactly to the lower-star filtration when $\varepsilon=0$. Moreover, for~every simplex $\sigma\in K$,
\[
\lim_{p\to\infty} f_p(\sigma)
=
\max_{v\in\sigma} f(v),
\]
and
\[
\lim_{\beta\to\infty} f_\beta(\sigma)
=
\max_{v\in\sigma} f(v).
\]

Thus, the~lower-star filtration is recovered exactly from the $\varepsilon$-Lower-Star method when $\varepsilon=0$, and~is obtained as the limiting case of the $L_p$ and Softmax methods as $p\to\infty$ and $\beta\to\infty$, respectively.
\end{remark}

While closely related to LSF, the~proposed filtrations provide systematic rules for assigning filtration values rather than modifying selected simplices individually. If~prior knowledge identifies a particular simplex that should enter later than its faces, its filtration value can be adjusted directly, or~the complex can be augmented before applying LSF. However, such simplices may not be known in advance. The~proposed methods therefore provide general rules that can be applied to the entire complex without requiring prior identification of particular simplices. Furthermore, LpF and SMF incorporate information from all vertices of a simplex rather than relying solely on the maximum vertex~value.

Each of the proposed constructions satisfies the monotonicity condition required for a valid filtration. Specifically, for~any simplices $\tau,\sigma\in K$ with $\tau\subseteq\sigma$,$f_*(\tau)\leq f_*(\sigma),$ where $f_*$ denotes any of $f_{\varepsilon}$, $f_p$, or~$f_{\beta}$. The~following proposition establishes this property and hence shows that each proposed construction defines a valid~filtration.

\begin{proposition}[Monotonicity]
Let \(K\) be a simplicial complex with vertex set \(V\), and~let
$f:V\to\mathbb{R}_{\geq 0}$ be a nonnegative real-valued function defined on the vertices. Then, for~any simplices $\tau,\sigma\in K$ with $\tau\subseteq\sigma$, $f_{\varepsilon}(\tau)\leq f_{\varepsilon}(\sigma),$
$f_p(\tau)\leq f_p(\sigma),$ 
$f_{\beta}(\tau)\leq f_{\beta}(\sigma).$
Therefore, each of the proposed constructions satisfies the required monotonicity condition and hence defines a valid filtration of \(K\).
\end{proposition}

\begin{proof}

\noindent\textbf{The \(\boldsymbol{\varepsilon}\)-Lower-Star Filtration.}

Let \(\tau,\sigma\in K\) satisfy $\tau\subseteq\sigma$.
Since every vertex of \(\tau\) is also a vertex of \(\sigma\),
\[
\max_{v\in\tau}f(v)
\leq
\max_{v\in\sigma}f(v).
\]

Moreover,
\[
\dim(\tau)\leq\dim(\sigma).
\]

Since \(\varepsilon\geq0\),
\[
\begin{aligned}
f_{\varepsilon}(\tau)
&=
\max_{v\in\tau}f(v)
+
\varepsilon\dim(\tau)\\
&\leq
\max_{v\in\sigma}f(v)
+
\varepsilon\dim(\sigma)\\
&=
f_{\varepsilon}(\sigma).
\end{aligned}
\]

Hence, $\tau\subseteq\sigma
\quad\Longrightarrow\quad
f_{\varepsilon}(\tau)\leq f_{\varepsilon}(\sigma),$ so \(f_{\varepsilon}\) defines a valid~filtration.

\textbf{The \(\boldsymbol{L_p}\) Filtration.} 

Let \(\tau,\sigma\in K\) satisfy $\tau\subseteq\sigma$.
Since \(f(v)\geq0\) for every \(v\in V\),
\[
\sum_{v\in\tau}f(v)^p
\leq
\sum_{v\in\sigma}f(v)^p.
\]

Because the function \(x\mapsto x^{1/p}\) is increasing on \([0,\infty)\),
\[
\begin{aligned}
f_p(\tau)
&=
\left(
\sum_{v\in\tau}f(v)^p
\right)^{1/p}\\
&\leq
\left(
\sum_{v\in\sigma}f(v)^p
\right)^{1/p}\\
&=
f_p(\sigma).
\end{aligned}
\]

Hence, $\tau\subseteq\sigma
\quad\Longrightarrow\quad
f_p(\tau)\leq f_p(\sigma),$ so \(f_p\) defines a valid~filtration.

\textbf{The Softmax Filtration.}

Let \(\tau,\sigma\in K\) satisfy $\tau\subseteq\sigma$.
Since \(\tau\subseteq\sigma\) and \(e^{\beta f(v)}>0\) for every \(v\in V\),
\[
\sum_{v\in\tau}e^{\beta f(v)}
\leq
\sum_{v\in\sigma}e^{\beta f(v)}.
\]

Because the logarithm is increasing and \(\beta>0\),
\[
\begin{aligned}
f_{\beta}(\tau)
&=
\frac{1}{\beta}
\log\left(
\sum_{v\in\tau}e^{\beta f(v)}
\right)\\
&\leq
\frac{1}{\beta}
\log\left(
\sum_{v\in\sigma}e^{\beta f(v)}
\right)\\
&=
f_{\beta}(\sigma).
\end{aligned}
\]

Hence, $\tau\subseteq\sigma
\quad\Longrightarrow\quad
f_{\beta}(\tau)\leq f_{\beta}(\sigma),$ so \(f_{\beta}\) defines a valid~filtration.

Therefore, each of the proposed constructions satisfies the required monotonicity condition and hence defines a valid filtration of \(K\).

\end{proof}

Example \ref{ex1} illustrates how the proposed filtration methods assign filtration values to the simplices of a finite graph and demonstrates the resulting \mbox{filtration~sequences.}

\begin{example} \label{ex1}: Consider a graph $G=(V,E)$ with node set as 
$V=\{v_1,v_2,v_3,v_4,v_5\}$ and edge set as 
$E=\{(v_1,v_2), (v_2,v_3), (v_3,v_4), (v_4,v_1), (v_2,v_5), (v_3,v_5), (v_2,v_4)\},$ as shown in Figure~\ref{fig}. The~graph $G$ can be viewed as a one-dimensional simplicial complex consisting of five $0$-simplices (vertices) and seven $1$-simplices (edges). By~constructing the corresponding clique (flag) complex, the~three $2$-simplices (filled triangles) $\{v_1,v_2,v_4\}$, $\{v_2,v_3,v_4\}$, and~$\{v_2,v_3,v_5\}$ are introduced, yielding a two-dimensional simplicial complex. This toy example is used to illustrate how the proposed filtrations are~constructed.

To define the filtration, we assign a real-valued function $f:V \to \mathbb{R}_{\geq 0}$ to the vertices as follows:$\quad f(v_1)=1,  \quad f(v_2)=4, \quad f(v_3)=2,\quad f(v_4)=5, \quad f(v_5)=6$
\begin{figure}[H]
 \centering
    \includegraphics[width=0.12\linewidth]{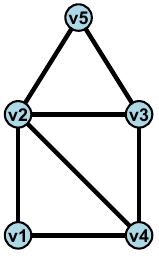}
    \caption{A toy graph $G$ with five vertices and seven edges, together with three filled triangles $(v_2,v_5,v_3)$, $(v_1,v_2,v_4)$, and~$(v_2,v_4,v_3)$, forming a two-dimensional simplicial complex with five $0$-simplices, seven $1$-simplices, and~three $2$-simplices.}
    \label{fig}
\end{figure}
\end{example}

\begin{enumerate}
\item \noindent \textbf{The Lower-star Filtration (LSF):} For the lower-star filtration, $\quad$
\[f_{\max}(\sigma)=\max_{v\in\sigma} f(v).\]
\end{enumerate}

Under this construction, each simplex receives a filtration value determined by the maximum function value among its vertices. Since vertices are \(0\)-simplices, their filtration values remain unchanged: $f_{\max}(v_1)=1,
f_{\max}(v_2)=4,
f_{\max}(v_3)=2,
f_{\max}(v_4)=5,
f_{\max}(v_5)=6.$ Edges and higher-dimensional simplices receive filtration values in the same manner. For~example, an~edge such as \((v_1,v_2)\) receives the filtration value $f_{\max}(v_1,v_2)=\max(1,4)=4$ and therefore enters the filtration at $t=4$. Similarly, the~triangle \((v_1,v_2,v_4)\) receives the filtration value $f_{\max}(v_1,v_2,v_4)=\max(1,4,5)=5,$ and enters the filtration at $t=5$

Applying this construction to the simplicial complex shown in Figure~\ref{fig} yields the following filtration~sequence:
\begin{itemize}
    \item $t=1$: $K_1=\{v_1\}$
    \item $t=2$: $K_2=K_1\cup\{v_3\}$
    
    \item $t=4$: $K_4=K_2\cup\{v_2,(v_1,v_2),(v_2,v_3)\}$
    
    \item $t=5$: $K_5=K_4\cup\{v_4,(v_1,v_4),(v_2,v_4),(v_3,v_4),
    (v_1,v_2,v_4),(v_2,v_3,v_4)\}$
    \item $t=6$: $K_6=K_5\cup\{v_5,(v_2,v_5),(v_3,v_5),
    (v_2,v_3,v_5)\}$
\end{itemize}

\begin{figure}[H]
    \centering
    \includegraphics[width=1.0\linewidth]{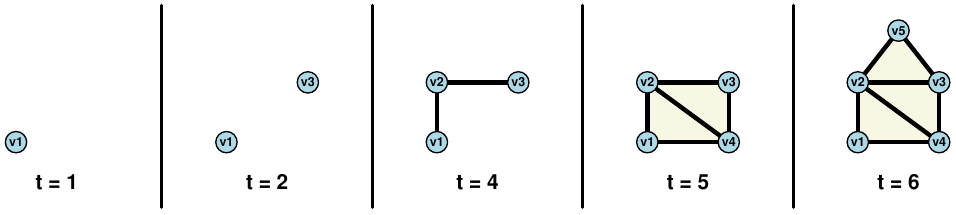}
   \caption{Geometric evolution of the lower-star filtration across increasing filtration thresholds. Simplices are added according to the maximum function value among their~vertices.}
    \label{fig1}
\end{figure}

As illustrated in Figure~\ref{fig1}, the~simplicial complex grows progressively as the filtration threshold increases. However, because~higher-dimensional simplices inherit the maximum value of their vertices, many simplices enter simultaneously with their lower-dimensional faces. For~example, the~triangles \((v_1,v_2,v_4)\) and \((v_2,v_3,v_4)\) enter the filtration at \(t=5\), the~same filtration value at which the vertex \(v_4\) and several associated edges enter the filtration. As~a result, loops are filled at the same filtration value at which they are formed, yielding \(H_1\) features with zero~persistence.

\begin{enumerate}
    \setcounter{enumi}{1}
\item \noindent \textbf{\(\boldsymbol{\varepsilon}\)-Lower-star Filtration (ELSF):} For the \(\varepsilon\)-lower-star filtration, let \(\varepsilon=0.1\). Then
\[
f_{\varepsilon}(\sigma)
=
\max_{v\in\sigma} f(v)
+
0.1\cdot\dim(\sigma).
\]
\end{enumerate}

Under this construction, each simplex receives a filtration value determined by the maximum function value among its vertices together with a dimension-dependent offset. Since vertices are \(0\)-simplices, their filtration values remain unchanged:
$f_{0.1}(v_1)=1,
f_{0.1}(v_2)=4,
f_{0.1}(v_3)=2,
f_{0.1}(v_4)=5,
f_{0.1}(v_5)=6.$
Edges and higher-dimensional simplices receive filtration values by adding a dimension-dependent offset to the maximum vertex value within the simplex. For~example, an~edge such as \((v_1,v_2)\) receives the filtration value $f_{0.1}(v_1,v_2)=\max(1,4)+0.1 (1) =4.1,$ and therefore enters the filtration at $t=4.1$.
Similarly, the~triangle \((v_1,v_2,v_4)\) receives the filtration value
$f_{0.1}(v_1,v_2,v_4)=\max(1,4,5)+0.1 (2) =5.2,$ and enters the filtration at $t=5.2$

The resulting filtration sequence~is:
\begin{itemize}
    \item $t=1$: $K_1=\{v_1\}$
    \item $t=2$: $K_2=K_1\cup\{v_3\}$
    \item $t=4$: $K_4=K_2\cup\{v_2\}$
    \item $t=4.1$: $K_{4.1}=K_4\cup\{(v_1,v_2),(v_2,v_3)\}$
    \item $t=5$: $K_5=K_{4.1}\cup\{v_4\}$
    \item $t=5.1$: $K_{5.1}=K_5\cup\{(v_1,v_4),(v_2,v_4),(v_3,v_4)\}$
    \item $t=5.2$: $K_{5.2}=K_{5.1}\cup\{(v_1,v_2,v_4),(v_2,v_3,v_4)\}$
    \item $t=6$: $K_6=K_{5.2}\cup\{v_5\}$
    \item $t=6.1$: $K_{6.1}=K_6\cup\{(v_2,v_5),(v_3,v_5)\}$
    \item $t=6.2$: $K_{6.2}=K_{6.1}\cup\{(v_2,v_3,v_5)\}$
\end{itemize}

\begin{figure}[H]
\centering
    \includegraphics[width=1.0\linewidth]{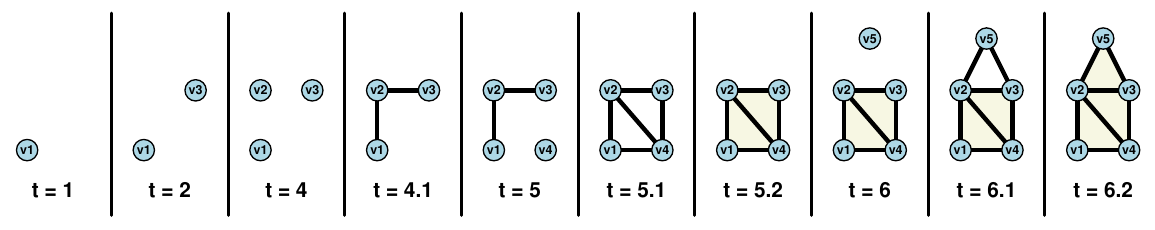}
    \caption{Geometric evolution of the \(\varepsilon\)-lower-star filtration across increasing filtration~thresholds.}
    \label{fig2}
\end{figure}

As illustrated in Figure~\ref{fig2}, the~introduction of the \(\varepsilon\)-offset separates the entry of higher-dimensional simplices from their boundary faces, producing a more gradual evolution of the simplicial complex. Consequently, edges and triangles no longer enter the filtration simultaneously, allowing loops to persist briefly before being filled. This reduces the number of simplices entering at the same filtration value and produces a more refined filtration structure. For~example, the~edges \((v_1,v_2)\), \((v_1,v_4)\), and~\((v_2,v_4)\) enter the filtration by \(t=5.1\), forming a \(1\)-dimensional loop. However, the~corresponding triangle \((v_1,v_2,v_4)\) does not enter the filtration until \(t=5.2\). As~a result, the~loop persists over a nonzero interval before being filled, allowing the associated \(H_1\) feature to be captured more effectively by persistent~homology.

\begin{enumerate}
    \setcounter{enumi}{2}
\item \noindent \textbf{The \(\boldsymbol{L_p}\)-Filtration (LpF):} For the \(L_p\)-filtration, let \(p=1\). Then \[f_{1}(\sigma)=\sum_{v\in\sigma} f(v).\]
\end{enumerate}

Under this construction, each simplex is assigned the sum of the function values of its vertices. Since vertices are \(0\)-simplices, their filtration values remain unchanged: $f_{1}(v_1)=1,
f_{1}(v_2)=4,
f_{1}(v_3)=2,
f_{1}(v_4)=5,
f_{1}(v_5)=6.$
Edges and higher-dimensional simplices receive filtration values by summing the function values of all vertices within the simplex. For~example, an~edge such as \((v_1,v_2)\) receives the filtration value \(f_{1}(v_1,v_2)=1+4=5\)
, while the triangle \((v_1,v_2,v_4)\) receives the filtration value
\(f_{1}(v_1,v_2,v_4)=1+4+5=10.
\)

The resulting filtration sequence~is:
\begin{itemize}
    \item $t=1$: $K_1=\{v_1\}$
    \item $t=2$: $K_2=K_1\cup\{v_3\}$
    \item $t=4$: $K_4=K_2\cup\{v_2\}$
    \item $t=5$: $K_5=K_4\cup\{v_4,(v_1,v_2)\}$
    \item $t=6$: $K_6=K_5\cup\{v_5,(v_2,v_3),(v_1,v_4)\}$
    \item $t=7$: $K_7=K_6\cup\{(v_3,v_4)\}$
    \item $t=8$: $K_8=K_7\cup\{(v_3,v_5)\}$
    \item $t=9$: $K_9=K_8\cup\{(v_2,v_4)\}$
    \item $t=10$: $K_{10}=K_9\cup\{(v_2,v_5),(v_1,v_2,v_4)\}$
    \item $t=11$: $K_{11}=K_{10}\cup\{(v_2,v_3,v_4)\}$
    \item $t=12$: $K_{12}=K_{11}\cup\{(v_2,v_3,v_5)\}$
\end{itemize}

\begin{figure}[H]
    \centering
    \includegraphics[width=1.0\linewidth]{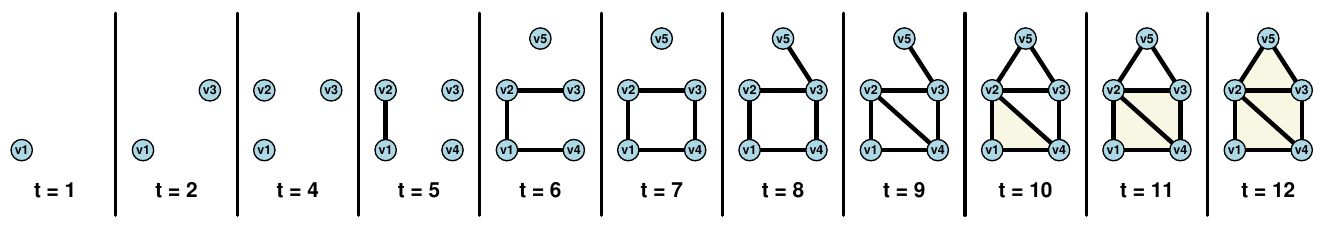}
    \caption{Geometric evolution of the \(L_1\)-filtration across increasing filtration thresholds. Simplices are added according to the sum of the function values of all vertices within each~simplex.}
    \label{fig3}
\end{figure}

As illustrated in Figure~\ref{fig3}, the~\(L_1\)-filtration produces a richer ordering of simplices by incorporating information from all vertices rather than relying only on the maximum vertex value. Consequently, simplices that would otherwise share the same filtration value become more separated, leading to a more detailed evolution of the simplicial complex. This additional separation reduces the number of simplices entering simultaneously and provides a clearer representation of the underlying topological structure. For~example, the~edges \((v_1,v_2)\), \((v_1,v_4)\), and~\((v_2,v_4)\) enter the filtration at \(t=5\), \(t=6\), and~\(t=9\), respectively, forming a \(1\)-dimensional hole at \(t=9\). However, the~corresponding triangle \((v_1,v_2,v_4)\) does not enter the filtration until \(t=10\). As~a result, the~loop persists over a nonzero interval before being filled, allowing the associated \(H_1\) feature to be captured more effectively by persistent~homology.

\begin{enumerate}
    \setcounter{enumi}{3}
\item \noindent \textbf{The Softmax Filtration (SMF):} For the softmax filtration, let \(\beta=1\). Then
\[
f_{1}(\sigma)
=
\log\left(\sum_{v\in\sigma} e^{f(v)}\right).
\]
\end{enumerate}

Under this construction, each simplex is assigned a filtration value through a softmax aggregation of the function values of its vertices. Since each vertex is a singleton simplex, \(f_1(v_i)=\log(e^{f(v_i)})=f(v_i),\) the~vertex filtration values remain unchanged:
\(f_{1}(v_1)=1,
f_{1}(v_2)=4,
f_{1}(v_3)=2,
f_{1}(v_4)=5,
f_{1}(v_5)=6.\)

Edges and higher-dimensional simplices receive filtration values through the softmax aggregation of all vertex values within the simplex. For~example, an~edge such as \((v_1,v_2)\) receives the filtration value
\(f_{1}(v_1,v_2)
=
\log\left(e^1+e^4\right)
\approx 4.05,\)
while the triangle \((v_1,v_2,v_4)\), receives the filtration value
\(f_{1}(v_1,v_2,v_4)
=
\log\left(e^1+e^4+e^5\right)
\approx 5.33.
\)

The resulting filtration sequence~is:
\begin{itemize}
    \item $t=1$: $K_{1}=\{v_1\}$
    \item $t=2$: $K_{2}=K_{1}\cup\{v_3\}$
    \item $t=4$: $K_{4}=K_{2}\cup\{v_2\}$
    \item $t=4.05$: $K_{4.05}=K_{4}\cup\{(v_1,v_2)\}$
    \item $t=4.13$: $K_{4.13}=K_{4.05}\cup\{(v_2,v_3)\}$
    \item $t=5$: $K_{5}=K_{4.13}\cup\{v_4\}$
    \item $t=5.02$: $K_{5.02}=K_{5}\cup\{(v_1,v_4)\}$
    \item $t=5.05$: $K_{5.05}=K_{5.02}\cup\{(v_3,v_4)\}$
    \item $t=5.31$: $K_{5.31}=K_{5.05}\cup\{(v_2,v_4\}$
     \item $t=5.33$: $K_{5.33}= K_{5.31}\cup\{(v_1,v_2,v_4)\}$
      \item $t=5.35$: $K_{5.35}= K_{5.33}\cup\{(v_2,v_3,v_4)\}$
       \item $t=6$: $K_{6}= K_{5.35}\cup\{v_5\}$
    \item $t=6.02$: $K_{6.02}=K_{6}\cup\{(v_3,v_5)\}$
\item $t=6.13$: $K_{6.13}=K_{6.02}\cup\{(v_2,v_5)\}$
    \item $t=6.14$: $K_{6.14}=K_{6.13}\cup\{(v_2,v_3,v_5)\}$
\end{itemize}

\begin{figure}[H]
    
    \includegraphics[width=1.0\linewidth]{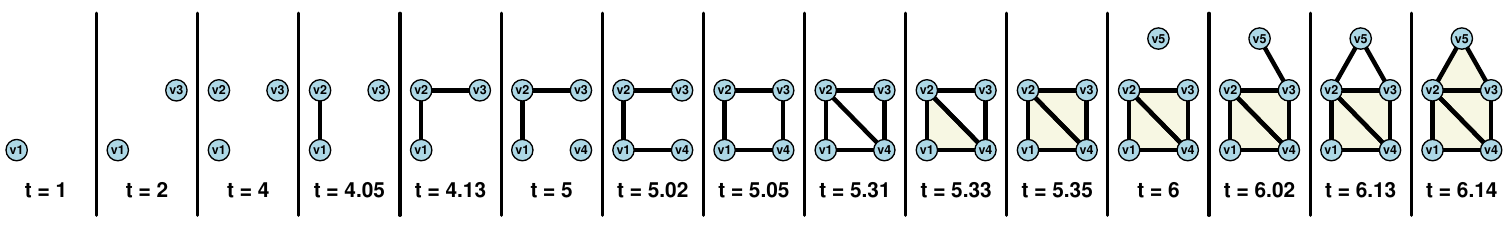}
    \caption{Geometric evolution of the softmax filtration across increasing filtration thresholds. Simplices are added using a softmax aggregation of the vertex function~values.}
    \label{fig4}
\end{figure}

As illustrated in Figure~\ref{fig4}, the~softmax filtration provides a smooth alternative to the lower-star filtration by incorporating weighted contributions from all vertices within a simplex rather than relying solely on the maximum vertex value. Consequently, simplices that would otherwise share identical filtration values become more separated, producing a smoother evolution of the simplicial complex and a more refined filtration structure. For~example, the~edges \((v_1,v_2)\), \((v_1,v_4)\), and~\((v_2,v_4)\) enter the filtration at \(t=4.05\), \(t=5.02\), and~\(t=5.31\), respectively, forming a \(1\)-dimensional hole at \(t=5.31\). However, the~corresponding triangle \((v_1,v_2,v_4)\) does not enter until \(t=5.33\). As~a result, the~loop persists briefly before being filled, allowing the associated \(H_1\) feature to be captured more effectively by persistent~homology.

\begin{figure}[H]
    \includegraphics[width=1.0\linewidth]{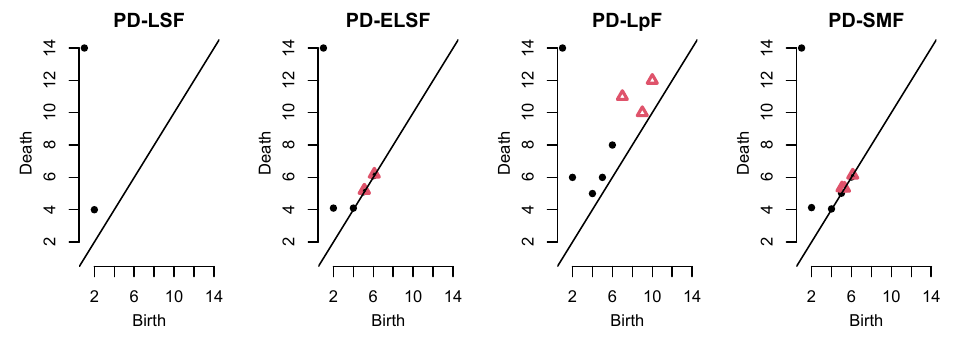}
    \caption{Persistence
 diagrams corresponding to the lower-star filtration (LSF), \(\varepsilon\)-lower-star filtration (ELSF), \(L_p\)-filtration (LpF), and~softmax filtration (SMF).}
    \label{fig5}
\end{figure}

Figure~\ref{fig5} presents the persistence diagrams obtained using the lower-star filtration (LSF), \(\varepsilon\)-lower-star filtration (ELSF), \(L_p\)-filtration (LpF), and~softmax filtration (SMF) for the toy graph \(G\) (see Example~\ref{ex1}). Compared with LSF, the~proposed filtrations produce additional \(H_1\) features and greater separation of persistence points from the diagonal. In~particular, LpF exhibits the largest separation, indicating longer persistence of the corresponding loop features. This behavior reflects how the proposed filtrations separate simplex entry times, allowing loops to persist across a wider range of filtration values before being filled. In~contrast, for~the toy graph considered here, LSF produces no \(H_1\) features because higher-dimensional simplices enter simultaneously with their boundary faces, causing the loops to be filled immediately upon~formation.

\unskip


\subsection{Stability of the Proposed Filtration~Methods}
\label{sec:stability}

Stability is a fundamental requirement in persistent homology, since small perturbations of the function defining a filtration should not produce large changes in the resulting persistence diagrams. The~stability theorem for persistence diagrams~\cite{cohen2007stability} formalizes this principle by showing that, for~continuous tame functions $f$ and $g$ defined on a triangulable space $X$,
\[
d_B\!\left(D(f),D(g)\right)
\leq
\|f-g\|_{\infty},
\]
where $D(f)$ and $D(g)$ denote the persistence diagrams associated with $f$ and $g$, respectively, and~$d_B$ denotes the bottleneck~distance.

Let $K$ be an abstract simplicial complex with vertex set $V$, and~$X = |K|$ its geometric realization. For~monotone functions $f, g: K \to \mathbb{R}$ extended to continuous tame functions on $X$, the~above stability theorem implies that
\begin{equation}\label{stability_bound}
d_B(D_k(f), D_k(g)) \leq \max_{\{\sigma\in K | \dim(\sigma)\leq k+1\}}|f(\sigma) - g(\sigma)|,
\end{equation}
where $D_k(f)$ and $D_k(g)$ are $k$-dimensional the persistence diagrams of filtrations induced by functions $f$ and $g$ respectively. It suffices to take the maximum over simplices of dimension at most $k+1$, since simplices of dimension $k+2$ or higher do not affect the $k$-dimensional diagrams. In~the following theorem we use this result to derive stability bounds for the proposed $\varepsilon$-Lower-Star, $L_p$, and~Softmax filtration~functions.

\begin{theorem}
Let $K$ be a finite simplicial complex with vertex set $V$, and~let $f,g:V\rightarrow\mathbb{R}_{\geq0}$ be two nonnegative vertex functions. Let $\varepsilon\geq0$, $1\leq p<\infty$, and~$\beta>0$. Then for every homological dimension $k$,
\begin{align}
   d_B\!\left(D_k(f_{\varepsilon}),D_k(g_{\varepsilon})\right) &\leq
\|f-g\|_{\infty,V},\\
d_B\!\left(D_k(f_p),D_k(g_p)\right) &\leq (k+2)^{1/p}\|f-g\|_{\infty,V},\\
d_B\!\left(D_k(f_{\beta}),D_k(g_{\beta})\right)
&\leq
\|f-g\|_{\infty,V},
\end{align}
where $D_k(\cdot)$ denotes the $k$-dimensional persistence diagram and $\|f-g\|_{\infty,V}=\max_{v\in V}|f(v)-g(v)|$.

\end{theorem}

\begin{proof}
We establish the stability bound for each filtration method~separately.

\textbf{The $\boldsymbol{\varepsilon}$-Lower-Star Filtration.}
For every simplex $\sigma\in K$,
\[
\begin{aligned}
\left|
f_{\varepsilon}(\sigma)-g_{\varepsilon}(\sigma)
\right|
&=
\left|
\max_{v\in\sigma}f(v)
+\varepsilon\dim(\sigma)
-\max_{v\in\sigma}g(v)
-\varepsilon\dim(\sigma)
\right|\\
&=
\left|
\max_{v\in\sigma}f(v)
-\max_{v\in\sigma}g(v)
\right|\leq
\max_{v\in\sigma}|f(v)-g(v)|\\
&=
\|f-g\|_{\infty,V}.
\end{aligned}
\]

Taking the maximum over all simplices of dimension at most $k+1$ and applying the stability bound (\ref{stability_bound}) to $f_{\varepsilon}$ and $g_{\varepsilon}$ gives

\[
\begin{aligned}
d_B\!\left(
D_k(f_{\varepsilon}),
D_k(g_{\varepsilon})
\right)
&\leq
\max_{\{\sigma\in K | \dim(\sigma)\leq k+1\}}
\left|
f_{\varepsilon}(\sigma)-g_{\varepsilon}(\sigma)
\right|\leq
\|f-g\|_{\infty,V}.
\end{aligned}
\]

\textbf{The $\boldsymbol{L_p}$ Filtration.}
By the reverse triangle inequality for the $L_p$-norm
\begin{equation}\label{inverse_triangle}
 \left|
f_p(\sigma)-g_p(\sigma)
\right|
\leq
\left(
\sum_{v\in\sigma}
|f(v)-g(v)|^p
\right)^{1/p}\leq
\left(
\sum_{v\in\sigma}
\|f-g\|_{\infty,V}^{p}
\right)^{1/p}=
|\sigma|^{1/p}\|f-g\|_{\infty,V}.   
\end{equation}

Since simplices of dimension $k+2$ or higher do not create or destroy $k$-dimensional topological features, it suffices to consider simplices of dimension at most $k+1$, \mbox{and~consequently}
\begin{equation} \label{sigma_bound}
 |\sigma|^{1/p}\leq(k+2)^{1/p}.   
\end{equation}

Inequalities (\ref{inverse_triangle}) and (\ref{sigma_bound}) imply that
\[
d_B\!\left(
D_k(f_p),
D_k(g_p)
\right)
\leq
\max_{\{\sigma\in K | \dim(\sigma)\leq k+1\}}
|f_p(\sigma)-g_p(\sigma)|\leq
(k+2)^{1/p}\|f-g\|_{\infty,V}.
\]

\textbf{The Softmax Filtration.}
Observe that for every $v\in V$,
\[
f(v)\leq g(v)+\|f-g\|_{\infty,V}.
\]

Multiplying by $\beta>0$ and exponentiating both sides, and~then summing over $v\in\sigma$ gives
\[
\sum_{v\in\sigma}e^{\beta f(v)}
\leq
e^{\beta\|f-g\|_{\infty,V}}
\sum_{v\in\sigma}e^{\beta g(v)},
\]
and hence
\[
\begin{aligned}
f_{\beta}(\sigma)-g_{\beta}(\sigma)&=\frac{1}{\beta}\log
\left(
\sum_{v\in\sigma}e^{\beta f(v)}
\right)-\frac{1}{\beta}\log
\left(
\sum_{v\in\sigma}e^{\beta g(v)}
\right)=
\frac{1}{\beta}
\log
\left(
\frac{
\sum_{v\in\sigma}e^{\beta f(v)}
}{
\sum_{v\in\sigma}e^{\beta g(v)}
}
\right)\\
&\leq
\frac{1}{\beta}
\log
\left(
\frac{
e^{\beta\|f-g\|_{\infty,V}}
\sum_{v\in\sigma}e^{\beta g(v)}
}{
\sum_{v\in\sigma}e^{\beta g(v)}
}
\right)=
\|f-g\|_{\infty,V}.
\end{aligned}
\]

Similarly, using 
\[
g(v)\leq f(v)+\|f-g\|_{\infty,V}, \hbox{ for all } v\in\sigma,
\]
one can show that
\[
g_{\beta}(\sigma)-f_{\beta}(\sigma)\leq\|f-g\|_{\infty,V},
\]
which gives
\[
|f_{\beta}(\sigma)-g_{\beta}(\sigma)|\leq\|f-g\|_{\infty,V}.
\]

Again, applying the stability bound (\ref{stability_bound}) to $f_{\beta}$ and $g_{\beta}$, we get
\[
d_B\!\left(
D_k(f_{\beta}),
D_k(g_{\beta})
\right)
\leq
\max_{\{\sigma\in K | \dim(\sigma)\leq k+1\}}
|f_{\beta}(\sigma)-g_{\beta}(\sigma)|\leq
\|f-g\|_{\infty,V}.
\]
\end{proof}

To complement the theoretical stability results, numerical evaluations of the derived stability bounds are provided in Appendix~\ref{app:stability}. The~evaluations consider different parameter choices for ELSF, LpF, and~SMF using \emph{degree} and \emph{betweenness} centrality as filter functions. As~shown in Tables~\ref{tab:stability_degree} and~\ref{tab:stability_betweenness}, the~observed bottleneck distances remain below the corresponding theoretical bounds for all parameter values~considered.

\section{Hypothesis Testing Procedure for Persistence~Diagrams}
\label{sec4} 

This section reviews the hypothesis testing framework proposed by \citet{robinson2017hypothesis} for comparing collections of persistence diagrams. Consider two independent samples of diagrams, $Z_1 = \{D_i\}_{i=1}^{n_1}$ and $Z_2 = \{\tilde{D}_i\}_{i=1}^{n_2}$, generated from data such as point clouds, images, or~graphs using an appropriate filtration. The~objective is to determine whether the two samples arise from the same underlying population or data-generating process. Formally, this problem is expressed through the hypotheses
\[
H_0: \mathcal{L}(D) = \mathcal{L}(\tilde{D}), \qquad
H_1: \mathcal{L}(D) \neq \mathcal{L}(\tilde{D}),
\]
where $\mathcal{L}(D)$ and $\mathcal{L}(\tilde{D})$ denote the laws governing the respective data-generating processes. The~null hypothesis states that both samples are generated from the same underlying process, while the alternative hypothesis indicates that the underlying processes~differ. 

Testing these hypotheses is difficult because persistence diagrams reside in a highly structured, non-Euclidean space, making standard parametric methods  difficult to apply. \citet{robinson2017hypothesis} therefore proposed a nonparametric permutation test based on a joint loss function that aggregates within-group distances. The~test statistic is defined as
\begin{equation}\label{eq:hypo}
L(Z_1,Z_2)
=
\frac{1}{2n_1(n_1-1)}
\sum_{i=1}^{n_1}\sum_{j=1}^{n_1}
d(D_i,D_j)
+
\frac{1}{2n_2(n_2-1)}
\sum_{i=1}^{n_2}\sum_{j=1}^{n_2}
d(\widetilde{D}_i,\widetilde{D}_j).
\end{equation}

Here, $d(\cdot,\cdot)$ denotes a distance metric, such as the bottleneck or Wasserstein distance. This statistic measures the average pairwise dissimilarity within each sample. Under~the null hypothesis, the~persistence diagrams are exchangeable, meaning that their assignment to either sample is statistically arbitrary. To~determine statistical significance, the~null distribution is generated empirically by repeatedly permuting the sample labels and recomputing $L(Z_1,Z_2)$ for each permutation. For~a significance level $\alpha$, the~null hypothesis is rejected if the observed statistic falls within the lower $\alpha$ tail of the permutation-based reference~distribution.

Although this framework provides a principled,  nonparametric approach for comparing persistence diagrams, it has two primary characteristics that motivate our modifications. First, the~test statistic focuses exclusively on within-group dispersion and does not explicitly account for between-group separation. Second, computing pairwise bottleneck or Wasserstein distances can be computationally expensive for large datasets because these distances require finding optimal matchings between persistence diagrams. To~improve computational efficiency, following \citet{islambekov2024vector}, we replace bottleneck and Wasserstein distances with distances computed on vectorized summaries of persistence diagrams, thereby avoiding expensive matching-based computations. In~addition, rather than relying on the joint loss function in Equation~(\ref{eq:hypo}), we adopt the energy statistic (E-statistic) of \citet{szekely2013energy} as the test statistic, following a strategy similar to~\cite{you2021topics}. Unlike the original loss function, which primarily measures within-group variability, the~E-statistic incorporates both within-group and between-group distances, providing an alternative formulation for comparing the two~samples.

\subsection{Energy Statistic for Two-Sample~Testing}

To quantify differences between probability distributions, we use the energy statistic (E-statistic). Let $X$ and $Y$ be random vectors in $\mathbb{R}^m$ with distributions $F$ and $G$, respectively. The~population energy quantity based on a distance function $d$ is defined as
\begin{equation}
E_d(F,G)
=
2\,\mathbb{E}[d(X,Y)]
-
\mathbb{E}[d(X,X')]
-
\mathbb{E}[d(Y,Y')],
\end{equation}
where $X$ and $X'$ are independent random vectors with distribution $F$, $Y$ and $Y'$ are independent random vectors with distribution $G$, and~all four random vectors are mutually independent. The~expectations are assumed to be finite. The~first term measures the expected distance between observations from different distributions, while the remaining terms measure the expected within-distribution~distances.

To apply the energy statistic in our setting, each persistence diagram is first transformed into a vectorized summary using an appropriate summary function (see Section~\ref{sub}). This representation allows persistence diagrams to be treated as observations in a vector space. Let $\beta_i$ denote the vectorized summary corresponding to the persistence diagram $D_i$, and~let $\tilde{\beta}_j$ denote the vectorized summary corresponding to $\tilde{D}_j$. The~resulting summaries form two independent samples, with~\[
\{\beta_i\}_{i=1}^{n_1} \overset{\mathrm{i.i.d.}}{\sim} F,
\qquad
\{\tilde{\beta}_j\}_{j=1}^{n_2} \overset{\mathrm{i.i.d.}}{\sim} G.
\]
The hypothesis testing problem is therefore formulated in terms of the vectorized summaries as
\[
H_0:F=G,\qquad H_1:F\neq G,
\]
where $F$ and $G$ denote the distributions of the summaries from the two~groups.

The corresponding generalized energy distance of \citet{szekely2013energy}, based on the Euclidean norm, is
\begin{equation}
E_{\gamma,2}(F,G)
=
2\,\mathbb{E}\|X-Y\|_2^\gamma
-
\mathbb{E}\|X-X'\|_2^\gamma
-
\mathbb{E}\|Y-Y'\|_2^\gamma,
\qquad 0<\gamma<2,
\end{equation}
where $\|\cdot\|_2$ denotes the Euclidean norm and $\gamma$ is the energy exponent. Under~appropriate moment conditions, $E_{\gamma,2}(F,G)$ is nonnegative and equals zero if and only if $F=G$. Consequently, larger values of $E_{\gamma,2}(F,G)$ indicate greater separation between the two distributions. When $F=G$, the~expected between-group distance is balanced by the expected within-group distances, yielding $E_{\gamma,2}(F,G)=0$. In~contrast, when $F\neq G$, the~population Euclidean energy distance is~positive.

In our proposed testing procedure, we use the $L_1$ norm with energy exponent $\gamma=1$ instead of the Euclidean norm. Replacing $\|\cdot\|_2$ with $\|\cdot\|_1$ in the preceding formulation yields $E_{1,1}(F,G)$. Since the $L_1$ norm is the sum of coordinate-wise absolute differences, $E_{1,1}(F,G)$ is equivalent to the sum of the corresponding univariate energy distances. Consequently, $E_{1,1}(F,G)=0$ if and only if the corresponding marginal distributions of $F$ and $G$ are identical, which does not necessarily imply equality of their joint~distributions.

Since the underlying distributions $F$ and $G$ are unknown, the~population quantity $E_{1,1}(F,G)$ is estimated from the observed samples. The~corresponding empirical energy statistic is
\begin{equation}
\begin{aligned}
\mathcal{E}_{1,1}
&=
\frac{2}{n_1n_2}
\sum_{i=1}^{n_1}\sum_{j=1}^{n_2}
\|\beta_i-\tilde{\beta}_j\|_1
-
\frac{1}{n_1^2}
\sum_{i=1}^{n_1}\sum_{j=1}^{n_1}
\|\beta_i-\beta_j\|_1
-
\frac{1}{n_2^2}
\sum_{i=1}^{n_2}\sum_{j=1}^{n_2}
\|\tilde{\beta}_i-\tilde{\beta}_j\|_1.
\end{aligned}
\label{eq:energy_empirical}
\end{equation}

To test the null hypothesis $H_0:F=G$, we use the scaled empirical energy statistic
\begin{equation}
T_{1,1}
=
\frac{n_1n_2}{n_1+n_2}\,
\mathcal{E}_{1,1},
\label{eq:energy_test}
\end{equation}
which serves as the test statistic for comparing the two samples. Under~$H_0$, the~population quantity $E_{1,1}(F,G)$ is zero, and~the between-group and within-group distances are expected to be comparable. Under~alternatives involving differences in the corresponding marginal distributions, $E_{1,1}(F,G)$ is positive, and~larger observed values of the test statistic provide evidence against the null hypothesis. An important property of the scaled energy statistic in Equation~(\ref{eq:energy_test}) is its relationship with the within-group loss in Equation~(\ref{eq:hypo}). When the two groups have equal sample sizes ($n_1=n_2$) and the same
pairwise distances are used, the two statistics are affine transformations of one another with a negative slope. Consequently, they yield identical permutation $p$-values when the same permutations are used. However, when the group sizes are unequal ($n_1\neq n_2$), this relationship does not generally hold, and the two statistics may produce different permutation $p$-values.

Because the sampling distribution of $T_{1,1}$ under the null hypothesis is generally unknown, we employ a nonparametric permutation procedure to approximate it. Let $Z=\{\beta_1,\ldots,\beta_{n_1},\tilde{\beta}_1,\ldots,\tilde{\beta}_{n_2}\}$ denote the pooled sample of vectorized summaries. Under~$H_0$, these observations are exchangeable, meaning that any partition of $Z$ into two~groups of sizes $n_1$ and $n_2$ is equally likely. To~approximate the null distribution, we repeatedly permute the group labels and recompute the test statistic, producing the permutation statistics $\{T_{1,1}^{*(1)},\ldots,T_{1,1}^{*(K)}\}$. The~observed test statistic, denoted by $T_{1,1}^{\mathrm{obs}}$, is then compared with this permutation distribution to compute the permutation $p$-value. The~validity of the permutation test under the null hypothesis follows from exchangeability and does not require $E_{1,1}(F,G)$ to characterize equality of arbitrary joint distributions. Thus, the~permutation test remains valid under $H_0:F=G$, although~its ability to distinguish alternative distributions is limited by the marginal characterization described above. The~significance level of the test is denoted by $\alpha$. The~complete permutation procedure is summarized in Algorithm~\ref{alg:energy_perm_test}.

\begin{algorithm}[H]
\caption{Permutation Test for the Energy~Statistic}
\label{alg:energy_perm_test}
\begin{algorithmic}

\State \textbf{Input:} Samples $\{\beta_i\}_{i=1}^{n_1}$ and $\{\tilde{\beta}_j\}_{j=1}^{n_2}$, number of permutations $K$, and~significance level $\alpha$.

\State Form the pooled sample $Z=\{\beta_1,\dots,\beta_{n_1},\tilde{\beta}_1,\dots,\tilde{\beta}_{n_2}\}$.

\State Compute the observed test statistic $T_{1,1}^{\mathrm{obs}}$ using Equation~\eqref{eq:energy_test}.

\State Initialize $count \leftarrow 0$.

\For{$k=1$ \textbf{to} $K$}

    \State Randomly permute the group labels of the observations in $Z$.

    \State Partition the permuted sample into two groups, $Z_1^{*(k)}$ and $Z_2^{*(k)}$, of~sizes $n_1$ and $n_2$.

    \State Compute the permuted test statistic $T_{1,1}^{*(k)}$.

    \If{$T_{1,1}^{*(k)} \ge T_{1,1}^{\mathrm{obs}}$}
        \State $count \leftarrow count + 1$
    \EndIf

\EndFor

\State \textbf{Output:} Empirical $p$-value $=\frac{count+1}{K+1}$.

\end{algorithmic}
\end{algorithm}

\subsection{Energy Statistic for Multiple-Sample~Testing}

The two-sample framework extends naturally to the $k$-sample setting, where the goal is to determine whether $k\geq 3$ independent samples arise from a common underlying distribution. Formally, we test the null hypothesis $H_0:F_1=F_2=\cdots=F_k$ against the alternative hypothesis that at least two distributions~differ.

Suppose we observe $k$ independent groups of vectorized summaries, with~\[
\beta_i^{(g)}\overset{\mathrm{i.i.d.}}{\sim}F_g,
\qquad i=1,\ldots,n_g,\quad g=1,\ldots,k.
\]

The multi-sample energy statistic extends the two-sample energy statistic by comparing all pairs of groups and aggregating the resulting pairwise energy quantities into a single global measure of distributional separation. Using the $L_1$ norm with energy exponent $\gamma=1$, the~population mutual energy quantity is defined as
\begin{equation}
E_{1,1}(F_1,\dots,F_k)
=
\sum_{1\leq g<h\leq k}
E_{1,1}(F_g,F_h),
\end{equation}
where $E_{1,1}(F_g,F_h)$ denotes the two-sample population energy quantity between $F_g$ and $F_h$. This quantity is nonnegative and equals zero if and only if the corresponding coordinate-wise marginal distributions are identical across all groups. However, equality of these marginal distributions does not necessarily imply equality of the joint distributions, since the relationships between coordinates may differ across~groups.

Following the two-sample formulation, the~corresponding empirical multi-sample energy statistic is defined as
\begin{equation}
\begin{aligned}
T_{1,1}^{(k)}
&=
\sum_{1\leq g<h\leq k}
\frac{n_gn_h}{n_g+n_h}
\Bigg[
\frac{2}{n_gn_h}
\sum_{i=1}^{n_g}\sum_{j=1}^{n_h}
\|\beta_i^{(g)}-\beta_j^{(h)}\|_1
\\[2mm]
&\qquad
-
\frac{1}{n_g^2}
\sum_{i=1}^{n_g}\sum_{j=1}^{n_g}
\|\beta_i^{(g)}-\beta_j^{(g)}\|_1
-
\frac{1}{n_h^2}
\sum_{i=1}^{n_h}\sum_{j=1}^{n_h}
\|\beta_i^{(h)}-\beta_j^{(h)}\|_1
\Bigg].
\end{aligned}
\label{eq:energy_multisample}
\end{equation}

The first term measures the average distance between observations from different groups, while the remaining terms measure the corresponding within-group distances. The~statistic aggregates all $\binom{k}{2}$ pairwise group comparisons to provide a global measure of separation among the $k$ distributions. The~weights $\frac{n_gn_h}{n_g+n_h}$ account for the sample sizes of the groups involved in each pairwise~comparison.

As in the two-sample setting, the~sampling distribution of $T_{1,1}^{(k)}$ under the null hypothesis is generally unknown and is approximated using a nonparametric permutation procedure. Let $Z=\bigcup_{g=1}^{k}\{\beta_i^{(g)}\}_{i=1}^{n_g}$ denote the pooled sample of vectorized summaries. Under~$H_0$, the~observations are exchangeable. The~null distribution is therefore generated by repeatedly repartitioning $Z$ into $k$ groups of sizes $(n_1,\dots,n_k)$ and recomputing $T_{1,1}^{(k)}$. Consequently, Algorithm~\ref{alg:energy_perm_test} extends naturally to the $k$-sample setting by permuting labels across all $k$ groups and using $T_{1,1}^{(k)}$ as the test statistic. Under~the null hypothesis, exchangeability ensures the validity of the permutation~test.


\section[Experiments]{{Experiments}}
\label{sec5}

To evaluate the proposed filtration methods, we conduct a series of experiments examining their computational efficiency, discriminative power, sensitivity to parameter choice, and~performance on real-world data. In Section, we compare the runtime of LSF, ELSF, LpF, and~SMF for computing persistence diagrams as the number of graphs increases. In Section~\ref{RDPG}, we use simulated Dirichlet RDPG graphs to evaluate the empirical power of the proposed filtration methods for distinguishing graph populations with varying levels of structural separation. In Section~\ref{par}, we explore the effect of the tuning parameters associated with ELSF, LpF, and~SMF to assess the robustness of their empirical performance. Finally, in Sections~\ref{exp3} and~\ref{x-ray}, we apply the proposed methods to real-world graph datasets and chest X-ray images, respectively, to~evaluate their ability to distinguish structural differences in real data. {For all experiments involving the hypothesis testing procedure described in} Algorithm~\ref{alg:energy_perm_test}, {we use the $L_1$ norm with energy exponent $\gamma=1$.}

\subsection{Comparing the Runtime of Filtration~Methods}
\label{subsec:runtime_simulation}

This experiment compares the total run-time required by each filtration method (LSF, ELSF, LpF, and~SMF) for computing persistence diagrams as the dataset size increases using simulated graph data generated from the Erd\H{o}s--R\'enyi random graph model~\cite{erdos1959}. For~each setting, two classes of undirected graphs are generated from \(G(n,p_1)\) and \(G(n,p_2)\), where the number of nodes is fixed at \(n=100\), with~edge probabilities \(p_1=0.35\) and \(p_2=0.32\). For~each class, \(N\in\{50,100,150,\dots,500\}\) graphs are generated, yielding a total of \(2N\) graphs per dataset. Each graph is treated as a one-dimensional simplicial complex consisting of vertices and edges. To~extend this representation to include higher-dimensional simplices, we define real-valued functions on the vertices. Here, degree and betweenness centrality are used as filter functions to induce the filtrations. For~each filter function considered, a~separate filtration is~constructed.

For the proposed filtration methods, we consider the following parameter settings: \(\varepsilon=0.1\) for ELSF, \(p\in\{1,2,10\}\) for LpF, and~\(\beta\in\{1,2\}\) for SMF. Persistence diagrams are computed under each filter function, and~the total run-time is recorded. Each benchmark experiment is repeated 10 times using the \texttt{microbenchmark} package~\cite{microbenchmark} in \texttt{R}, and~the median run-time (in seconds) is reported. This procedure is repeated for each dataset size \(N\), allowing the scaling behavior of the computational cost to be examined as the number of graphs increases. The~results are summarized in Table~\ref{tab:runtime_comparison}.

\begin{table}[H]
\centering
\caption{Median runtime (seconds) for persistence diagram computation
across varying dataset sizes (total number of graphs $2N$).
The fastest method per row is indicated in blue, and the slowest in red.}
\label{tab:runtime_comparison}

\resizebox{\textwidth}{!}{%
\begin{tabular}{cccccccc}
\toprule
$\boldsymbol{2N}$ &
\textbf{LSF} &
\textbf{ELSF} $\boldsymbol{(\varepsilon=0.1)}$ &
\textbf{LpF} $\boldsymbol{(p=1)}$ &
\textbf{LpF} $\boldsymbol{(p=2)}$ &
\textbf{LpF} $\boldsymbol{(p=10)}$ &
\textbf{SMF} $\boldsymbol{(\beta=1)}$ &
\textbf{SMF} $\boldsymbol{(\beta=2)}$ \\
\midrule
100  & \textcolor{red}{46.25} & 45.89 & 24.81 & \textcolor{blue}{24.77} & 25.06 & 25.09 & 26.46 \\
200  & \textcolor{red}{88.24} & 87.40 & 48.47 & 39.56 & \textcolor{blue}{39.39} & 40.02 & 41.64 \\
300  & \textcolor{red}{108.35} & 106.84 & \textcolor{blue}{58.35} & \textcolor{blue}{58.35} & 59.98 & 61.14 & 63.47 \\
400  & \textcolor{red}{139.65} & 137.19 & \textcolor{blue}{78.23} & 78.52 & 80.96 & 81.54 & 85.13 \\
500  & \textcolor{red}{181.02} & 178.33 & \textcolor{blue}{98.64} & 98.98 & 101.96 & 103.64 & 109.66 \\
600  & \textcolor{red}{247.23} & 210.43 & \textcolor{blue}{121.34} & 122.44 & 126.15 & 128.96 & 135.77 \\
700  & \textcolor{red}{255.43} & 250.49 & \textcolor{blue}{143.10} & 145.63 & 149.26 & 152.56 & 159.99 \\
800  & \textcolor{red}{297.85} & 291.08 & \textcolor{blue}{164.39} & 167.62 & 173.30 & 177.99 & 188.31 \\
900  & \textcolor{red}{358.08} & 352.02 & \textcolor{blue}{211.10} & 230.17 & 234.19 & 221.44 & 214.06 \\
1000 & \textcolor{red}{422.35} & 409.50 & 253.73 & \textcolor{blue}{222.67} & 228.20 & 233.02 & 244.97 \\
\bottomrule
\end{tabular}%
}
\end{table}

The results show that the computational cost increases with dataset size for all filtration methods, with~noticeable differences in run-time among the methods. LSF consistently exhibits the largest run-time across all dataset sizes, making it the most computationally expensive method in this experiment. Although~ELSF provides a modest reduction in run-time relative to LSF, its computational cost remains substantially higher than those of LpF and~SMF.

LpF demonstrates the strongest computational performance overall, with~its different parameter choices achieving the lowest run-times across the dataset sizes considered. SMF also exhibits considerably lower run-times than both LSF and ELSF, although~its computational cost is generally slightly higher than that of LpF. The~difference becomes particularly noticeable for larger datasets. For~example, when (2N=1000), LSF requires approximately (422.35) seconds, whereas LpF with (p=2) requires approximately (222.67) seconds. These results demonstrate that LpF and SMF maintain substantially lower computational times than LSF as the number of graphs~increases.

\subsection{Distinguishing Simulated Dirichlet RDPG~Graphs}\label{RDPG}

To evaluate the empirical power of the proposed filtration methods under the modified hypothesis testing framework, we simulate graph data using the Dirichlet Random Dot Product Graph (RDPG) model~\cite{young2007random,sussman2012consistent}. In~this framework, each node is assigned a latent position vector sampled from a Dirichlet distribution with parameter \(\boldsymbol{\alpha}\in\mathbb{R}^3\). For~any two nodes with latent positions \(\mathbf{x}=(x_1,x_2,x_3)\) and \(\mathbf{y}=(y_1,y_2,y_3)\), an~edge is formed independently with probability
\(P(\mathbf{x},\mathbf{y})=x_1y_1+x_2y_2+x_3y_3,
\) corresponding to the dot product of their latent vectors. This construction provides a flexible probabilistic framework for generating graph populations with controlled structural~differences.

Two groups of graphs are generated from different Dirichlet distributions. For~the first group, latent positions are sampled from
\(\boldsymbol{\alpha}_1=(1.5,1.5,1.5),
\)
while for the second group, latent positions are sampled from
\(\boldsymbol{\alpha}_2(r)=(1.5,1.5,1.5+r),
\)
where \(r\) controls the separation between the two graph populations. As~\(r\) increases, the~latent distributions become increasingly different, leading to stronger structural differences between the generated graphs. Following~\cite{islambekov2024vector}, we consider \(r=0.4(i-1)\), \(i=1,\dots,5\), yielding progressively more distinguishable graph~distributions.

\begin{figure}[H]
    \centering
    \includegraphics[width=1.0\linewidth]{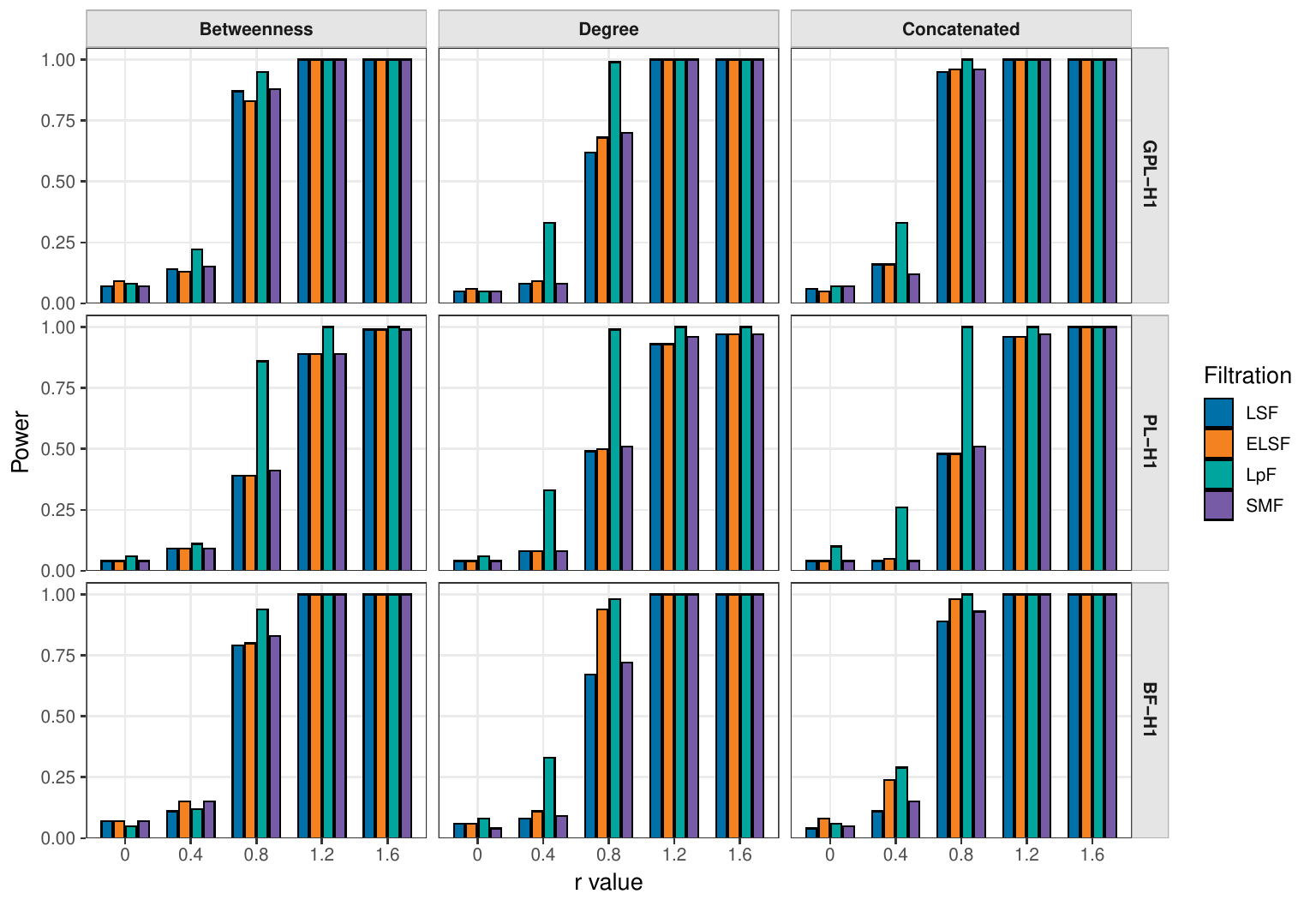}
    \caption{Empirical power of the modified energy-based permutation test in homological dimension \(H_1\) under simulated Dirichlet RDPG graph distributions for varying separation parameter \(r\). Results are shown for LSF, ELSF, LpF, and~SMF using degree and betweenness centrality as filter functions, together with their concatenated vector representations based on GPL, PL, and~BF.}
    \label{fig:RDPG_power}
\end{figure}

For each setting, \(n_1=n_2=10\) graphs with 100 nodes are generated for the two~groups. Degree and betweenness centrality are used as filter functions for constructing the filtrations. Unless~otherwise specified, we set \(\varepsilon=0.1\) for ELSF, \(p=1\) for LpF, and~\(\beta=1\) for SMF. Persistence diagrams are computed under each filter function and transformed into vector representations using PL, GPL, and~BF. Pairwise \(L_1\) distances between the resulting vector representations are used in the modified energy-based permutation test to assess differences between the graph populations. The~test is performed at the significance level \(\alpha=0.05\) using 999 permutations. Each experiment is repeated 100 times, and~empirical power is estimated as the proportion of times the null hypothesis is rejected. {The hypothesis tests are conducted separately for homological dimensions} \(H_0\), \(H_1\), and~\(H_2\), {using the same testing procedure across all filtration methods, filter functions, and~vector representations. The~results for} \(H_1\) are presented in
Figure~\ref{fig:RDPG_power}, {as this homological dimension demonstrates higher and more consistent statistical power across the filtration methods.The corresponding results for} \(H_0\) and \(H_2\) are provided in Appendix~\ref{app:rdpg_results}.

The results show that the empirical power of all filtration methods increases as the separation parameter \(r\) increases. However, noticeable differences are observed among the filtration methods, particularly for moderate values of \(r\). LpF and SMF generally outperform LSF across most settings, indicating greater sensitivity to structural differences between the graph~populations.

Among the proposed methods, LpF consistently demonstrates the strongest performance, often achieving higher empirical power at smaller values of \(r\) and reaching power close to one earlier than the competing methods. SMF also exhibits competitive performance and generally performs better than LSF, although~its performance is slightly less consistent than that of LpF in some settings. ELSF provides modest improvements over LSF, but~its performance is generally weaker than those of LpF and~SMF.

\subsubsection[Effect of Filtration Parameter Choice]{{Effect of Filtration Parameter Choice}} 
\label{par}
The proposed filtration methods depend on tuning parameters that control how vertex information is aggregated when assigning filtration values to simplices. It is therefore important to assess how the choice of these parameters affects the empirical power of the resulting topological~test.

To investigate this effect, we repeat the Dirichlet RDPG simulation described above while varying the tuning parameter associated with each proposed filtration method. Specifically, for~ELSF, we consider
\(\varepsilon\in\{0.01,0.025,0.05,0.10,0.20\};
\)
for LpF, we consider
\(p\in\{1,1.5,2,4,8,16\};
\)
and for SMF, we consider
\(\beta\in\{0.5,1,2,5,10,20\}.
\)
LSF is included as a reference for~comparison.

As in the preceding experiment, degree and betweenness centrality are used as filter functions, and~empirical power is evaluated over the separation values \(r=0.4(i-1)\), \(i=1,\ldots,5\). For~each choice of the filtration parameter, persistence diagrams are computed from the simulated graphs and transformed using BF and PL. The~modified energy-based permutation test is then applied, and~empirical power is estimated as the proportion of times the null hypothesis is rejected over the repeated~simulations.

\begin{figure}[H]
    \centering
    \includegraphics[width=1.0\linewidth]{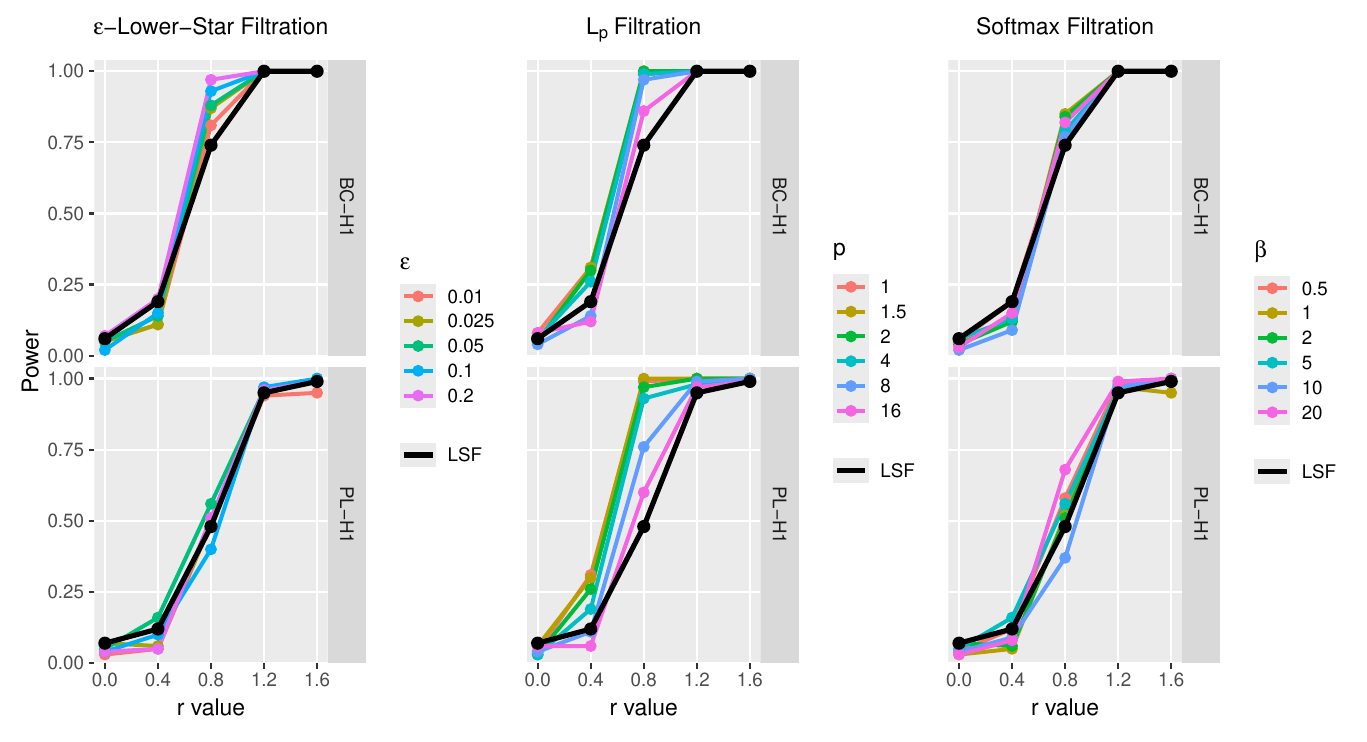}
    \caption{Empirical power of the modified energy-based permutation test for different filtration parameter choices using \emph{degree} centrality as the filter function. Results are shown for BF and PL summaries across varying separation parameter \(r\). The~columns correspond to the proposed filtration methods, and~the black curve represents~LSF.}
    \label{fig:parameter_degree}
\end{figure}

\begin{figure}[H]
\centering
    \includegraphics[width=1.0\linewidth]{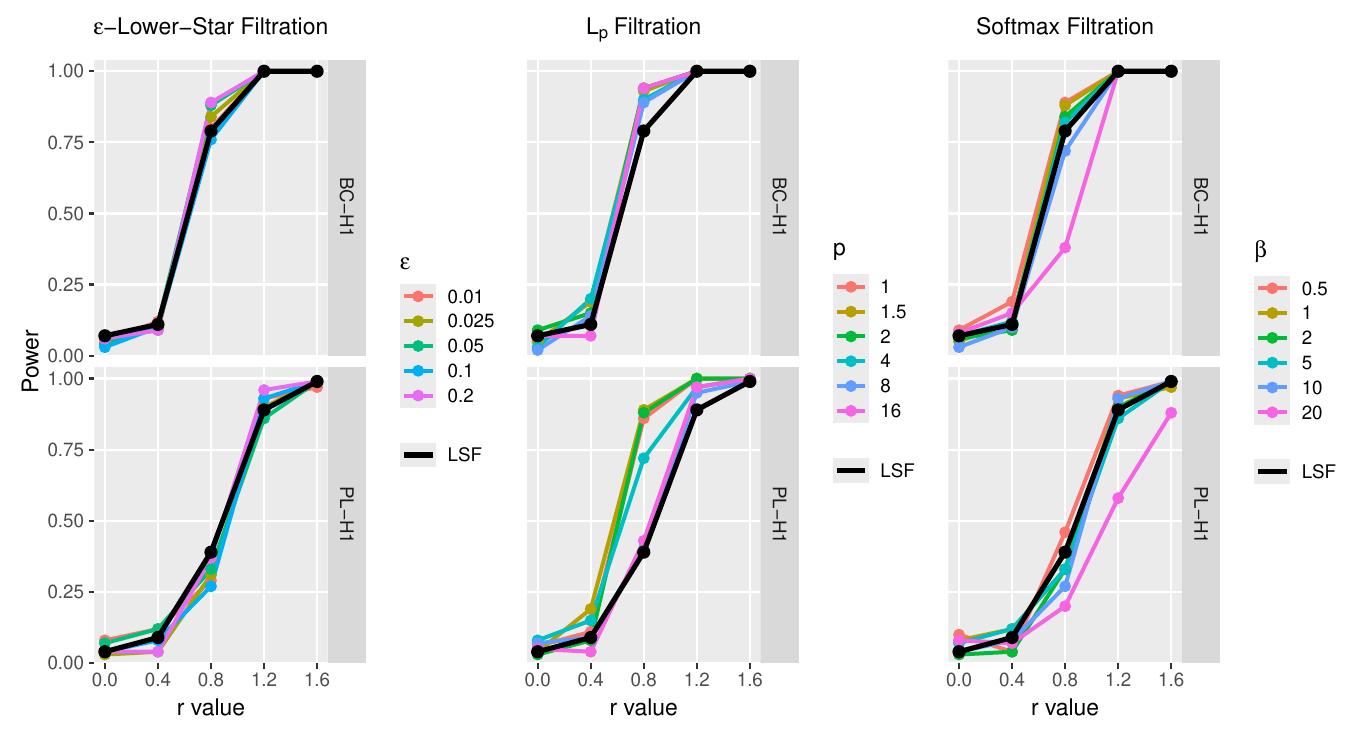}
    \caption{Empirical power of the modified energy-based permutation test for different filtration parameter choices using \emph{betweenness} centrality as the filter function. Results are shown for BF and PL summaries across varying values for separation parameter \(r\). 
    	The~columns correspond to the proposed filtration methods, and~the black curve represents~LSF.}
    \label{fig:parameter_betweenness}
\end{figure}

The resulting power curves are used to assess the sensitivity of each proposed filtration method to its tuning parameter and to compare its performance with LSF. Results are presented separately for each filter function using BF and PL, as~shown in Figures~\ref{fig:parameter_degree} and~\ref{fig:parameter_betweenness}. The~results indicate that the proposed filtration methods are generally robust to the choice of tuning parameter, with~similar empirical power patterns across the parameter values considered. Differences among parameter choices are most apparent at moderate values of \(r\), but~become small as the separation between the graph populations increases. Several parameter choices also yield higher empirical power than~LSF.

\unskip

\unskip

\subsection{Application 1: Distinguishing Real-World~Graphs}
\label{exp3}

The experiments in this section use datasets from TUDataset, a~widely used benchmark collection for graph classification and regression tasks~\cite{kersting2016benchmark}. We consider 10 datasets containing graphs classified into two or more groups (see Table~\ref{tab:real} in Appendix~\ref{app:datasets}). These datasets vary in the number of nodes, class sizes, and~degree of class imbalance, providing a diverse setting for evaluating the proposed filtration methods. Since topological features in homological dimension \(H_2\) were not consistently present across the datasets considered, the~empirical analysis focuses on homological dimensions \(H_0\) and \(H_1\).

For each graph, degree and betweenness centrality are used as filter functions to construct LSF, ELSF, LpF, and~SMF. We set \(\varepsilon=0.1\) for ELSF, \(p=1\) for LpF, and~\(\beta=1\) for SMF. The~resulting persistence diagrams are transformed into finite-dimensional vector representations using GPL, PL and BF. For~both GPL and PL, the~first \(k=3\) landscape functions are retained, and~each representation is evaluated on a fixed grid of 100 evenly spaced points spanning the range from the minimum birth value to the maximum death value observed across the sample. For~GPL, an~additional smoothing step is applied using an Epanechnikov kernel with bandwidth \(h=0.2\).

Pairwise \(L_1\) distances between the resulting vector representations are used as input to the modified energy-based permutation test (Algorithm~\ref{alg:energy_perm_test}) to assess whether graphs from different classes arise from distinct underlying distributions. The~test is performed using 999 permutations at the significance level \(\alpha=0.05\). For~each filtration method, results are obtained separately using degree and betweenness centrality and their concatenated vector~representations.

A related hypothesis testing framework based on LSF was previously studied in~\cite{islambekov2024vector}, and~we refer readers to this work for additional methodological details and comparison with our results. Here, the~modified energy-based permutation test is applied directly to the real-world graph datasets, and~the corresponding rejection and non-rejection outcomes are reported. The~results for \(H_0\) are summarized in Table~\ref{tab:results_estat_grouped}, while the corresponding \(H_1\) results are provided in Appendix~\ref{app:h1_results}.

The results show that the proposed filtration methods provide additional rejection outcomes relative to LSF across several datasets and topological summaries. The~differences are particularly apparent in settings where LSF fails to reject the null hypothesis but one or more of the proposed filtration methods detect a significant difference between the graph populations. LpF and SMF frequently produce rejection outcomes in such settings, while ELSF also provides additional rejections in several cases. These results suggest that the additional topological features produced by the proposed filtrations can provide useful information for detecting structural differences between graph~populations.

\begin{table}[H]
\centering
\caption{Summary of hypothesis testing results based on the energy
statistic for the real-world graph datasets in homological dimension
$H_0$. Results are reported for the three topological summary
functions (GPL, PL, and BF) across the four filtration methods
(LSF, ELSF, LpF, and SMF). Filtrations are induced using
betweenness and degree centrality as filter functions.
The columns $B_0$ and $D_0$ denote the $H_0$ topological
summaries obtained using betweenness and degree centrality,
respectively, while $C_0$ denotes their concatenated
representation. Test outcomes are denoted as follows:
\cmark\ indicates rejection of the null hypothesis ($p<0.05$),
\textcolor{blue}{*} indicates borderline evidence
($0.05 \leq p < 0.10$), and \xmark\ indicates failure to reject
the null hypothesis ($p \geq 0.10$).}
\label{tab:results_estat_grouped}

\resizebox{\textwidth}{!}{%
\begin{tabular}{l*{12}{c}}
\toprule
\multirow{2}{*}{\textbf{Dataset}}
& \multicolumn{3}{c}{\textbf{LSF}}
& \multicolumn{3}{c}{\textbf{ELSF}}
& \multicolumn{3}{c}{\textbf{LpF}}
& \multicolumn{3}{c}{\textbf{SMF}} \\
\cmidrule(lr){2-4}
\cmidrule(lr){5-7}
\cmidrule(lr){8-10}
\cmidrule(lr){11-13}
& $\boldsymbol{B_0}$ & $\boldsymbol{D_0}$ & $\boldsymbol{C_0}$
& $\boldsymbol{B_0}$ & $\boldsymbol{D_0}$ & $\boldsymbol{C_0}$
& $\boldsymbol{B_0}$ & $\boldsymbol{D_0}$ & $\boldsymbol{C_0}$
& $\boldsymbol{B_0}$ & $\boldsymbol{D_0}$ & $\boldsymbol{C_0}$ \\
\midrule

\multicolumn{13}{c}{\textbf{GPL-$H_0$}} \\
\midrule
DHFR
& \cmark & \cmark & \cmark
& \cmark & \cmark & \cmark
& \cmark & \cmark & \cmark
& \cmark & \cmark & \cmark \\

PROTEINS
& \cmark & \cmark & \cmark
& \cmark & \cmark & \cmark
& \xmark & \cmark & \cmark
& \cmark & \cmark & \cmark \\

IMDB-BINARY
& \cmark & \cmark & \cmark
& \xmark & \cmark & \cmark
& \cmark & \cmark & \cmark
& \cmark & \cmark & \cmark \\

REDDIT-BINARY
& \xmark & \xmark & \xmark
& \xmark & \cmark & \cmark
& \xmark & \xmark & \xmark
& \cmark & \cmark & \cmark \\

DD
& \xmark & \cmark & \cmark
& \cmark & \cmark & \cmark
& \xmark & \cmark & \cmark
& \cmark & \cmark & \cmark \\

COX2
& \cmark & \textcolor{blue}{*} & \cmark
& \cmark & \xmark & \cmark
& \cmark & \cmark & \cmark
& \cmark & \cmark & \cmark \\

BZR
& \cmark & \xmark & \cmark
& \cmark & \cmark & \cmark
& \cmark & \cmark & \cmark
& \cmark & \cmark & \cmark \\

ENZYMES
& \xmark & \cmark & \cmark
& \cmark & \cmark & \cmark
& \cmark & \cmark & \cmark
& \cmark & \cmark & \cmark \\

IMDB-MULTI
& \xmark & \cmark & \cmark
& \xmark & \cmark & \cmark
& \xmark & \cmark & \cmark
& \textcolor{blue}{*} & \cmark & \cmark \\

REDDIT-MULTI-5K
& \cmark & \cmark & \cmark
& \cmark & \cmark & \cmark
& \cmark & \cmark & \cmark
& \cmark & \cmark & \cmark \\
\midrule

\multicolumn{13}{c}{\textbf{PL-$H_0$}} \\
\midrule
DHFR
& \cmark & \cmark & \cmark
& \cmark & \cmark & \cmark
& \cmark & \cmark & \cmark
& \cmark & \cmark & \cmark \\

PROTEINS
& \cmark & \xmark & \cmark
& \cmark & \xmark & \cmark
& \cmark & \textcolor{blue}{*} & \cmark
& \cmark & \xmark & \cmark \\

IMDB-BINARY
& \cmark & \cmark & \cmark
& \cmark & \cmark & \cmark
& \cmark & \cmark & \cmark
& \cmark & \cmark & \cmark \\

REDDIT-BINARY
& \cmark & \cmark & \cmark
& \cmark & \cmark & \cmark
& \cmark & \cmark & \cmark
& \cmark & \cmark & \cmark \\

DD
& \xmark & \xmark & \xmark
& \xmark & \xmark & \xmark
& \xmark & \cmark & \xmark
& \cmark & \xmark & \cmark \\

COX2
& \cmark & \cmark & \cmark
& \cmark & \cmark & \cmark
& \cmark & \cmark & \cmark
& \cmark & \cmark & \cmark \\

BZR
& \cmark & \cmark & \cmark
& \cmark & \cmark & \cmark
& \cmark & \cmark & \cmark
& \cmark & \cmark & \cmark \\

ENZYMES
& \cmark & \xmark & \cmark
& \cmark & \xmark & \cmark
& \cmark & \textcolor{blue}{*} & \cmark
& \cmark & \xmark & \cmark \\

IMDB-MULTI
& \cmark & \cmark & \cmark
& \cmark & \cmark & \cmark
& \cmark & \cmark & \cmark
& \cmark & \cmark & \cmark \\

REDDIT-MULTI-5K
& \cmark & \cmark & \cmark
& \cmark & \cmark & \cmark
& \cmark & \cmark & \cmark
& \cmark & \cmark & \cmark \\
\midrule

\multicolumn{13}{c}{\textbf{BF-$H_0$}} \\
\midrule
DHFR
& \cmark & \cmark & \cmark
& \cmark & \cmark & \cmark
& \cmark & \cmark & \cmark
& \cmark & \cmark & \cmark \\

PROTEINS
& \cmark & \cmark & \cmark
& \cmark & \cmark & \cmark
& \cmark & \cmark & \cmark
& \cmark & \cmark & \cmark \\

IMDB-BINARY
& \cmark & \cmark & \cmark
& \cmark & \cmark & \cmark
& \cmark & \cmark & \cmark
& \cmark & \cmark & \cmark \\

REDDIT-BINARY
& \cmark & \cmark & \cmark
& \cmark & \cmark & \cmark
& \cmark & \cmark & \cmark
& \cmark & \cmark & \cmark \\

DD
& \cmark & \cmark & \cmark
& \cmark & \cmark & \cmark
& \cmark & \cmark & \cmark
& \cmark & \cmark & \cmark \\

COX2
& \cmark & \cmark & \cmark
& \cmark & \cmark & \cmark
& \cmark & \cmark & \cmark
& \cmark & \cmark & \cmark \\

BZR
& \cmark & \cmark & \cmark
& \cmark & \cmark & \cmark
& \cmark & \cmark & \cmark
& \cmark & \cmark & \cmark \\

ENZYMES
& \cmark & \xmark & \cmark
& \cmark & \textcolor{blue}{*} & \cmark
& \cmark & \cmark & \cmark
& \cmark & \cmark & \cmark \\

IMDB-MULTI
& \cmark & \cmark & \cmark
& \cmark & \cmark & \cmark
& \cmark & \cmark & \cmark
& \cmark & \cmark & \cmark \\

REDDIT-MULTI-5K
& \cmark & \cmark & \cmark
& \cmark & \cmark & \cmark
& \cmark & \cmark & \cmark
& \cmark & \cmark & \cmark \\
\bottomrule

\end{tabular}%
}
\end{table}

Compared with the framework studied in~\cite{islambekov2024vector}, which uses LSF together with degree and betweenness centrality as filter functions, the~modified energy-based permutation framework combined with the proposed filtration methods demonstrates improved ability to distinguish graph distributions for the BZR dataset. In~particular, the~proposed filtration methods detect distributional differences in nearly all settings in homological dimension \(H_0\) when using the PL and BF representations. In~contrast, the~framework in~\cite{islambekov2024vector} failed to reject the null hypothesis across the corresponding settings for this dataset. This comparison further illustrates how the proposed filtration methods, together with the modified testing framework, can provide additional sensitivity for distinguishing graph~populations.

\subsection{Application 2: Distinguishing Chest X-Ray~Images}
\label{x-ray}

\begin{figure}[H]
    \centering
    \includegraphics[width=0.83\linewidth]{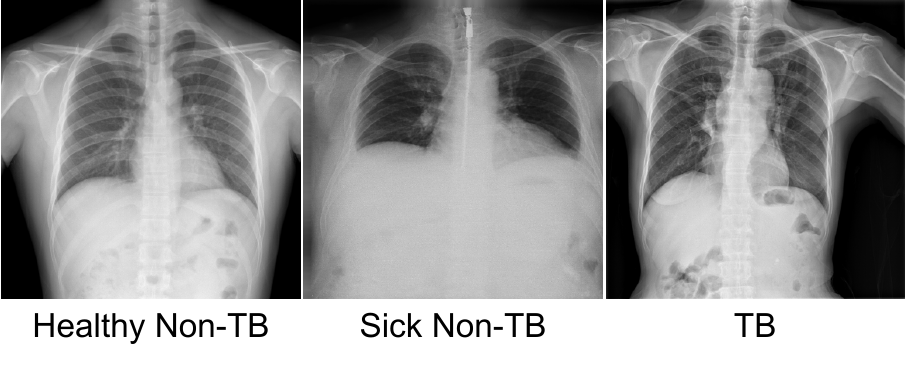}
    \caption{Representative chest X-ray images from the TBX11K dataset. From~left to right: healthy (normal), sick but non-TB, and~active TB~cases.}
    \label{fig:tbx_classes}
\end{figure}

In this experiment, we analyze chest X-ray images from the TBX11K (TBX-11) dataset~\cite{liu2020rethinking,chauhan2014role,jaeger2014two}, publicly available on Kaggle~\cite{tbx11k_kaggle}. The~dataset contains approximately 11,200 chest X-ray images with tuberculosis (TB) annotations and is designed for both classification and detection tasks. TBX11K is compiled from several publicly available chest X-ray datasets, including the Montgomery County and Shenzhen datasets~\cite{jaeger2014two}. The~images are organized into three categories: healthy (normal), sick but non-TB, and~active TB cases. Representative examples from these classes are shown in Figure~\ref{fig:tbx_classes}.

\begin{figure}[H]
\centering

\resizebox{0.9\textwidth}{!}{%
\begin{tikzpicture}


\node[font=\bfseries\large] at (2.5,6.5)
{Image as pixel grid};

\foreach \x/\y/\val/\shade in {
0/5/0.95/15, 1/5/0.75/28, 2/5/0.50/45, 3/5/0.25/65, 4/5/0.10/82,
0/4/0.90/18, 1/4/0.70/32, 2/4/0.45/50, 3/4/0.20/70, 4/4/0.08/85,
0/3/0.85/22, 1/3/0.65/35, 2/3/0.40/53, 3/3/0.18/73, 4/3/0.06/88,
0/2/0.80/25, 1/2/0.60/38, 2/2/0.35/57, 3/2/0.15/76, 4/2/0.04/90,
0/1/0.75/28, 1/1/0.55/42, 2/1/0.30/60, 3/1/0.10/80, 4/1/0.02/92
}{
    \fill[black!\shade]
    (\x,\y) rectangle ++(1,1);

    \draw[line width=0.8pt]
    (\x,\y) rectangle ++(1,1);

    \ifnum\shade>55
        \node[text=white,font=\small]
        at (\x+0.5,\y+0.5) {\val};
    \else
        \node[text=black,font=\small]
        at (\x+0.5,\y+0.5) {\val};
    \fi
}


\draw[->,ultra thick]
(5.35,3.5) -- (6.15,3.5);


\node[font=\bfseries\large] at (9.8,6.5)
{Pixels become vertices};

\foreach \x in {0,...,4}{
    \foreach \y in {0,...,4}{

        \pgfmathtruncatemacro{\lab}
        {5*(4-\y)+\x+1}

        \fill
        (7.2+\x,1+\y) circle (2.5pt);

        \node[font=\scriptsize]
        at (7.2+\x+0.15,1+\y+0.18)
        {$v_{\lab}$};
    }
}


\draw[->,ultra thick]
(12.55,3.5) -- (13.35,3.5);


\node[font=\bfseries\large] at (16.4,6.5)
{Triangulated simplicial complex};

\foreach \x in {14.4,15.4,16.4,17.4}{
    \foreach \y in {2,...,5}{

        \fill[blue!25,opacity=0.45]
        (\x,\y) --
        (\x+1,\y) --
        (\x+1,\y-1) -- cycle;

        \fill[blue!25,opacity=0.45]
        (\x,\y) --
        (\x,\y-1) --
        (\x+1,\y-1) -- cycle;
    }
}

\foreach \y in {1,...,5}{
    \draw[line width=0.9pt]
    (14.4,\y) -- (18.4,\y);
}

\foreach \x in {14.4,15.4,16.4,17.4,18.4}{
    \draw[line width=0.9pt]
    (\x,1) -- (\x,5);
}

\foreach \x in {14.4,15.4,16.4,17.4}{
    \foreach \y in {2,...,5}{

        \draw[line width=1pt]
        (\x,\y) -- (\x+1,\y-1);
    }
}

\foreach \x in {0,...,4}{
    \foreach \y in {0,...,4}{

        \pgfmathtruncatemacro{\lab}
        {5*(4-\y)+\x+1}

        \fill
        (14.4+\x,1+\y) circle (2.5pt);

        \node[font=\scriptsize]
        at (14.4+\x+0.15,1+\y+0.18)
        {$v_{\lab}$};
    }
}

\end{tikzpicture}%
}

\caption{Illustration of the image-to-simplicial-complex
construction using a $5\times5$ image grid.
The grayscale image patch is represented as a pixel grid,
where larger intensity values appear on the left and smaller
values appear on the right. Each pixel is treated as a vertex
in a lattice, neighboring pixels are connected, and each
square cell is subdivided into two triangles. This construction
produces a triangulated simplicial complex suitable for
filtrations induced by pixel intensities and subsequent
persistent homology analysis.}

\label{fig:image_to_simplicial_complex}

\end{figure}
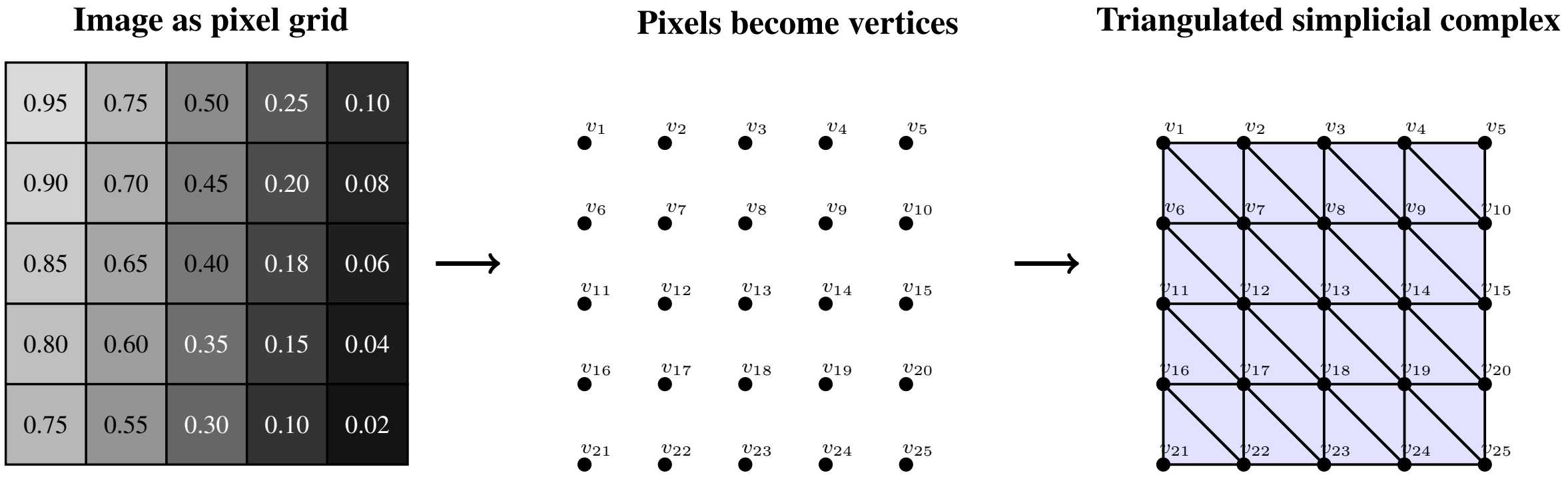

To apply the proposed filtration methods to image data, each image is resized to a resolution of \(256\times256\) pixels. Images with multiple color channels are converted to grayscale, and~any missing pixel values are replaced with zero. Each image is therefore represented as a matrix \(g\in[0,1]^{256\times256}\), where lower pixel values correspond to darker regions and higher values correspond to brighter regions. The~grayscale pixel intensities are then used as the filter function for constructing the~filtrations. Each processed image is represented as a simplicial complex by treating every pixel as a vertex in a lattice. To~incorporate spatial relationships among neighboring pixels, adjacent vertices are connected by edges, and~each \(2\times2\) cell is triangulated into two \(2\)-simplices. Repeating this construction across the image produces a triangulated simplicial complex on which persistent homology is computed. Figure~\ref{fig:image_to_simplicial_complex} illustrates this construction using a \(5\times5\) image~grid.

To extract topological features from the chest X-ray images, the~grayscale pixel intensities are used to induce LSF, ELSF, LpF, and~SMF on the resulting simplicial complexes. We set \(\varepsilon=0.1\) for ELSF, \(p=1\) for LpF, and~\(\beta=1\) for SMF. The~filtration methods differ in how the vertex values are aggregated to assign filtration values to higher-dimensional simplices. Consequently, they may produce persistence diagrams containing different topological features from the same~image.

The resulting persistence diagrams are transformed into finite-dimensional vector representations using GPL, PL, and~BF. For~both GPL and PL, the~first \(k=3\) landscape functions are retained, and~each representation is evaluated on a fixed grid of 100 evenly spaced points spanning the range from the minimum birth value to the maximum death value observed across the sample. For~GPL, an~additional smoothing step is applied using an Epanechnikov kernel with bandwidth \(h=0.2\).

To assess topological differences among the three image classes, we apply the modified energy-based permutation test. Pairwise \(L_1\) distances between the vectorized representations are computed and used as input to the test. The~null hypothesis states that the three image groups arise from the same underlying distribution and is rejected when the corresponding \(p\)-value is below the significance level \(\alpha=0.05\). The~test is performed using 999~permutations.

\begin{figure}[H]
    
    \includegraphics[width=1.0\linewidth]{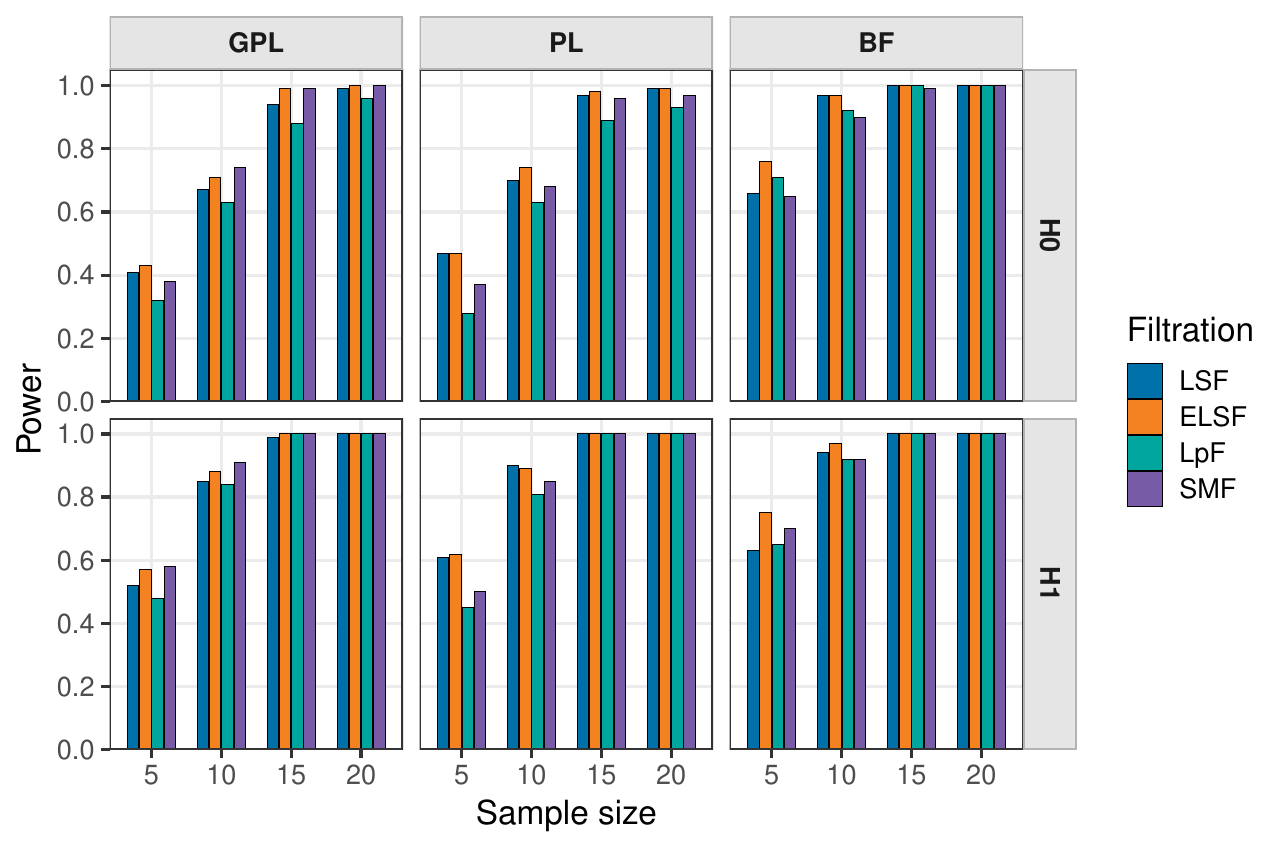}
    \caption{Estimated power of the modified energy-based permutation test for chest X-ray image data as sample size increases using images with resolution \(256\times256\) pixels at significance level \(\alpha=0.05\). Results are shown for four filtration methods (LSF, ELSF, LpF, and~SMF) and three topological summaries (GPL, PL, and~BF) across homological dimensions \(H_0\) (top row) and \(H_1\) (bottom row).}
    \label{fig13}
\end{figure}

To estimate empirical power, the~procedure is repeated over random subsamples of the dataset. For~each sample size \(n\in\{5,10,15,20\}\), \(n\) images are sampled without replacement from each class, resulting in \(3n\) images per experiment. This procedure is repeated over \(B=100\) Monte Carlo replications, and~empirical power is estimated as the proportion of replications in which the null hypothesis is rejected. Results are reported for homological dimensions 0 and 1, allowing the filtration methods to be compared in their ability to distinguish topological differences among the three chest X-ray image classes. 
{The analysis focuses on homological dimensions} \(H_0\) and \(H_1\), {because higher-dimensional homology is trivial for the planar simplicial complexes considered.} The results are presented in Figure~\ref{fig13}.

The results show that the proposed filtration methods generally achieve empirical power comparable to or higher than LSF, although~their relative performance varies across topological summaries and homological dimensions. ELSF exhibits the most consistent performance, frequently attaining the highest empirical power, particularly as the sample size increases. LpF and SMF also achieve higher empirical power than LSF in several settings, although~their relative performance depends on the topological summary and homological dimension~considered.

As the sample size increases, the~differences among the filtration methods become less pronounced, with~all methods approaching empirical power close to one. This indicates that the differences among filtration methods are most apparent at smaller and moderate sample sizes, where distinguishing the image classes is more challenging. The~results in Appendix~\ref{app:image_resolution} further show that increasing the image resolution from \(64\times64\) to \(128\times128\) and \(256\times256\) pixels generally increases empirical power across the topological summaries considered, with~the largest improvements occurring at smaller sample sizes. This suggests that higher image resolution can provide additional structural information that improves the ability of the proposed framework to distinguish the chest X-ray image~classes.


\section{Conclusions}\label{sec13}

This work introduced three alternative filtration methods:
the \(\varepsilon\)-lower-star filtration (ELSF), the~\(L_p\) {filtration (LpF), and~the softmax filtration (SMF).
These methods provide alternative approaches to constructing
filtrations and offer greater flexibility in capturing
topological features that may not be observed under the
lower-star filtration. The~stability results established
in this work provide theoretical support for the proposed
methods and demonstrate their robustness to perturbations
in the filter values.}

\textls[-15]{In addition, we extended the hypothesis testing framework
of~\mbox{\citet{robinson2017hypothesis}}} by adopting the energy
statistic of~\citet{szekely2013energy} and comparing
vectorized persistence diagram summaries using the
$L_1$ distance, following~\mbox{\citet{islambekov2024vector}.}
This formulation provides a computationally efficient
approach to comparing graph populations without requiring
pairwise bottleneck or Wasserstein distance~calculations.

Applications to real-world graph and image datasets
further demonstrate the effectiveness of the proposed
methods. The~experimental results show that the alternative
filtrations can improve sensitivity to structural
differences in the data while maintaining computational
efficiency.

The sensitivity analysis further shows that the proposed
filtration methods are generally robust across the
parameter values considered. While the present study does
not provide an automated procedure for selecting
\(\varepsilon\), \(p\), and~\(\beta\), the~results provide
an assessment of the influence of parameter choice within
the present hypothesis-testing framework. Automated
parameter selection may become particularly important
when the proposed filtration methods are used for
predictive tasks, such as classification and regression,
where parameter values can be selected based on predictive
performance. Developing data-driven parameter-selection
procedures for such applications represents an important
direction for future~work.


\appendix

\renewcommand{\thesection}{Appendix \Alph{section}}

\renewcommand{\thetable}{\Alph{section}\arabic{table}}
\renewcommand{\thefigure}{\Alph{section}\arabic{figure}}


\section{Numerical Evaluation of Stability Bounds}
\label{app:stability}

\setcounter{table}{0}
\setcounter{figure}{0}

To complement the theoretical stability results in Section~\ref{sec:stability}, we numerically evaluate the derived stability bounds for ELSF, LpF, and~SMF under different parameter choices. We use the Dirichlet RDPG model introduced in Section~\ref{RDPG} and generate a graph with \(100\) nodes using the baseline Dirichlet parameter $
\boldsymbol{\alpha}=(1.5,1.5,1.5).
$

Unlike the simulation study in Section~\ref{RDPG}, which compares two populations of graphs, the~present experiment examines the effect of perturbing the values of the filter function on a simulated graph. The~evaluations are performed using degree and betweenness centrality as filter functions. For~each filter function, let \(f\) denote the original filter function computed on the vertices of the graph. A~perturbed filter function \(g\) is constructed~as $
g(v)=\max\{0,f(v)+U_v\},
\quad
U_v\sim\operatorname{Uniform}(-\delta,\delta),
$

where $
\delta=
\frac{\max_v f(v)-\min_v f(v)}{10}.$ For both \(f\) and \(g\), the~clique complex \(K\) is constructed up to dimension \(2\), consisting of vertices, edges, and~triangles. The~filter values are then extended to the simplices using ELSF, LpF, and~SMF to obtain the corresponding filtrations. We consider
$\varepsilon\in{0,0.5,1,2,5},
p\in{1,2,4,8,16},
\beta\in{0.5,1,2,5,10},
$ for ELSF, LpF, and~SMF, respectively.

For each filtration method and parameter value, two persistence diagrams are computed: one from the filtration induced by \(f\) and the other from the filtration induced by \(g\). The~bottleneck distance, \(d_B\), between~the corresponding persistence diagrams is then calculated. We also compute the sup norm between the induced filtration functions, given by $
\max_{{\sigma\in K,|,\dim(\sigma)\leq k+1}}
\left|f_*(\sigma)-g_*(\sigma)\right|,
$ where \(f_*\) and \(g_*\) denote the corresponding ELSF, LpF, or~SMF filtration functions. This quantity is reported as the ``Sup Norm'' in \mbox{Tables~\ref{tab:stability_degree} and~\ref{tab:stability_betweenness}}. The~observed bottleneck distances and sup norms are then compared with the corresponding theoretical bounds established in Section~\ref{sec:stability}.

The results in Tables~\ref{tab:stability_degree} and~\ref{tab:stability_betweenness} show that the observed bottleneck distances remain below the corresponding sup norms, which in turn remain below the theoretical bounds, for~all filtration methods, homological dimensions, and~parameter values considered. These results provide numerical support for the stability results established in Section~\ref{sec:stability}.

\begin{table}[H]
\centering
\caption{Numerical stability results using degree centrality as the filter function.}
\label{tab:stability_degree}

\resizebox{\textwidth}{!}{%
\begin{tabular}{llcccccc}
\toprule
\multirow{2}{*}{\textbf{Method}}
& \multirow{2}{*}{\textbf{Parameter}}
& \multicolumn{3}{c}{$\boldsymbol{H_0}$}
& \multicolumn{3}{c}{$\boldsymbol{H_1}$} \\
\cmidrule(lr){3-5}
\cmidrule(lr){6-8}
&
& $\boldsymbol{d_B}$ & \textbf{Sup Norm} & \textbf{Bound}
& $\boldsymbol{d_B}$ & \textbf{Sup Norm} & \textbf{Bound} \\
\midrule

\multirow{5}{*}{ELSF}
& $\varepsilon=0$   & 1.828 & 2.791 & 2.791 & 2.727 & 2.791 & 2.791 \\
& $\varepsilon=0.5$ & 1.828 & 2.791 & 2.791 & 2.727 & 2.791 & 2.791 \\
& $\varepsilon=1$   & 1.828 & 2.791 & 2.791 & 2.727 & 2.791 & 2.791 \\
& $\varepsilon=2$   & 1.828 & 2.791 & 2.791 & 2.727 & 2.791 & 2.791 \\
& $\varepsilon=5$   & 2.281 & 2.791 & 2.791 & 2.727 & 2.791 & 2.791 \\
\midrule

\multirow{5}{*}{LpF}
& $p=1$  & 3.097 & 5.467 & 5.582 & 4.926 & 7.676 & 8.373 \\
& $p=2$  & 2.281 & 3.844 & 3.947 & 2.132 & 4.419 & 4.834 \\
& $p=4$  & 1.577 & 3.232 & 3.319 & 1.459 & 3.352 & 3.673 \\
& $p=8$  & 1.451 & 2.963 & 3.043 & 1.560 & 2.963 & 3.202 \\
& $p=16$ & 1.746 & 2.837 & 2.914 & 1.866 & 2.837 & 2.989 \\
\midrule

\multirow{5}{*}{SMF}
& $\beta=0.5$ & 1.683 & 2.791 & 2.791 & 1.858 & 2.791 & 2.791 \\
& $\beta=1$   & 1.817 & 2.791 & 2.791 & 2.416 & 2.791 & 2.791 \\
& $\beta=2$   & 1.828 & 2.791 & 2.791 & 2.664 & 2.791 & 2.791 \\
& $\beta=5$   & 1.828 & 2.791 & 2.791 & 2.726 & 2.791 & 2.791 \\
& $\beta=10$  & 1.828 & 2.791 & 2.791 & 2.727 & 2.791 & 2.791 \\
\bottomrule
\end{tabular}%
}
\end{table}

\begin{table}[H]
\centering
\caption{Numerical stability results using betweenness centrality as the filter function.}
\label{tab:stability_betweenness}

\resizebox{\textwidth}{!}{%
\begin{tabular}{llcccccc}
\toprule
\multirow{2}{*}{\textbf{Method}}
& \multirow{2}{*}{\textbf{Parameter}}
& \multicolumn{3}{c}{$\boldsymbol{H_0}$}
& \multicolumn{3}{c}{$\boldsymbol{H_1}$} \\
\cmidrule(lr){3-5}
\cmidrule(lr){6-8}
&
& $\boldsymbol{d_B}$ & \textbf{Sup Norm} & \textbf{Bound}
& $\boldsymbol{d_B}$ & \textbf{Sup Norm} & \textbf{Bound} \\
\midrule

\multirow{5}{*}{ELSF}
& $\varepsilon=0$   & 2.526 & 4.279 & 4.279 & 3.041 & 4.279 & 4.279 \\
& $\varepsilon=0.5$ & 2.526 & 4.279 & 4.279 & 3.041 & 4.279 & 4.279 \\
& $\varepsilon=1$   & 2.526 & 4.279 & 4.279 & 3.041 & 4.279 & 4.279 \\
& $\varepsilon=2$   & 2.526 & 4.279 & 4.279 & 3.041 & 4.279 & 4.279 \\
& $\varepsilon=5$   & 2.526 & 4.279 & 4.279 & 3.041 & 4.279 & 4.279 \\
\midrule

\multirow{5}{*}{LpF}
& $p=1$  & 6.420 & 8.271 & 8.558 & 7.687 & 12.166 & 12.837 \\
& $p=2$  & 3.033 & 5.811 & 6.052 & 2.576 & 6.776 & 7.412 \\
& $p=4$  & 2.526 & 4.846 & 5.089 & 3.168 & 5.085 & 5.632 \\
& $p=8$  & 2.526 & 4.426 & 4.666 & 3.053 & 4.459 & 4.909 \\
& $p=16$ & 2.526 & 4.289 & 4.469 & 3.041 & 4.289 & 4.583 \\
\midrule

\multirow{5}{*}{SMF}
& $\beta=0.5$ & 2.526 & 4.279 & 4.279 & 3.040 & 4.279 & 4.279 \\
& $\beta=1$   & 2.526 & 4.279 & 4.279 & 3.041 & 4.279 & 4.279 \\
& $\beta=2$   & 2.526 & 4.279 & 4.279 & 3.041 & 4.279 & 4.279 \\
& $\beta=5$   & 2.526 & 4.279 & 4.279 & 3.041 & 4.279 & 4.279 \\
& $\beta=10$  & 2.526 & 4.279 & 4.279 & 3.041 & 4.279 & 4.279 \\
\bottomrule
\end{tabular}%
}
\end{table}

\section[Additional Empirical Power Results for Dirichlet RDPG Simulations]{\texorpdfstring{\protect{Additional Empirical Power Results for Dirichlet RDPG~Simulations}}{Additional Empirical Power Results for Dirichlet RDPG Simulations}}
\label{app:rdpg_results}
This section presents the empirical power results for homological dimensions \(H_0\) and \(H_2\), complementing the \(H_1\) results reported in Section~\ref{RDPG}. The~results are obtained using the same simulation settings and hypothesis testing procedure described in Section~\ref{RDPG}, allowing a comparison of the filtration methods across all three homological~dimensions.

\begin{figure}[H]
    \centering
    \includegraphics[width=1.0\linewidth]{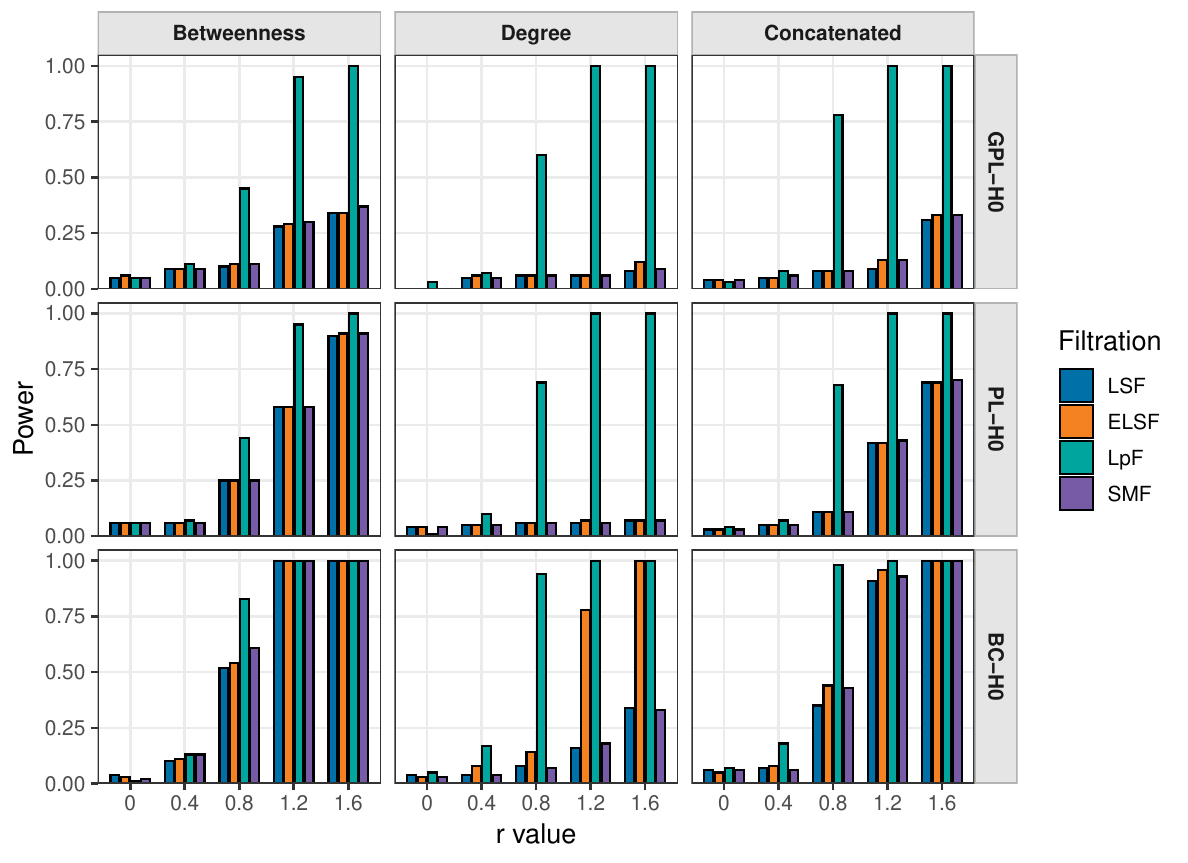}
    \caption{Empirical 
 power of the modified energy-based permutation test in homological dimension \(H_0\) under simulated Dirichlet RDPG graph distributions for varying separation parameter \(r\). Results are shown for LSF, ELSF, LpF, and~SMF using degree and betweenness centrality as filter functions, together with their concatenated vector representations based on GPL, PL, and~BF.}
    \label{fig:RDPG_power_H0}
\end{figure}
\unskip

\begin{figure}[H]
     \centering
    \includegraphics[width=1.0\linewidth]{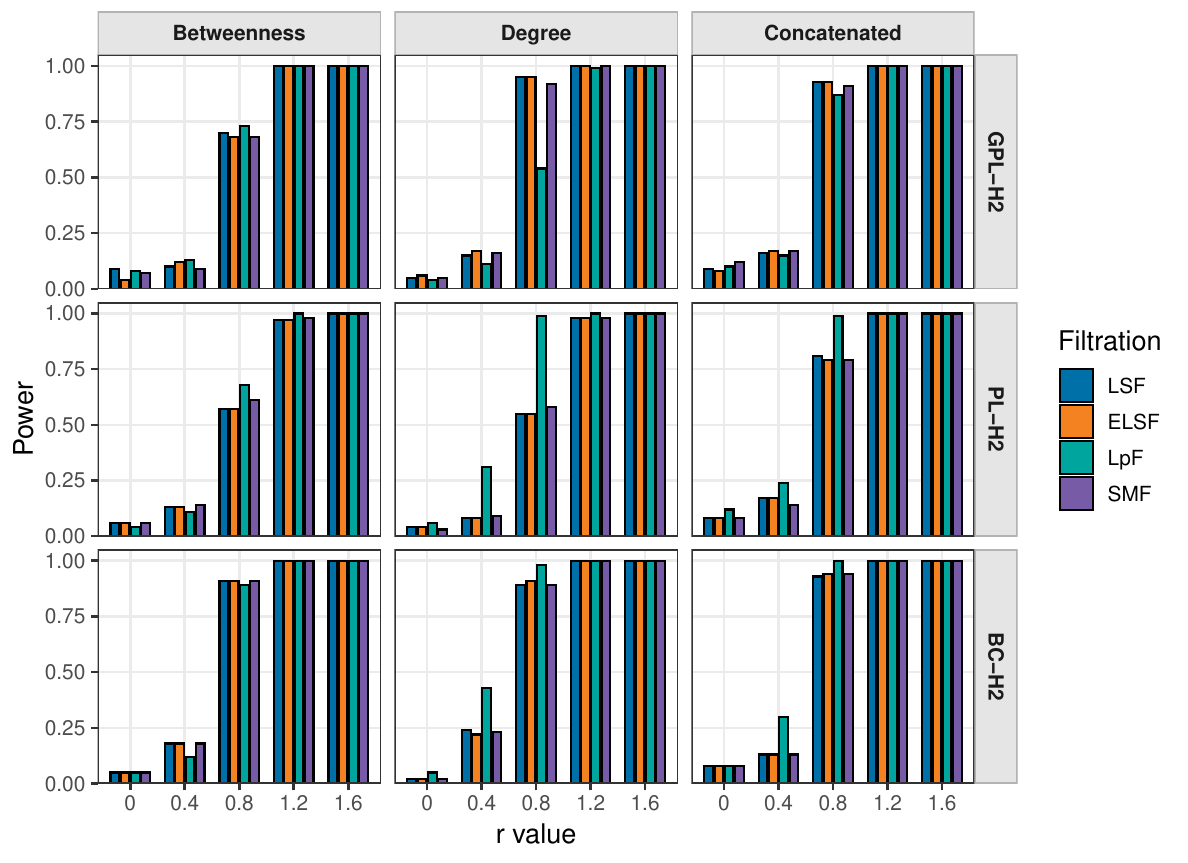}
    \caption{Empirical power of the modified energy-based permutation test in homological dimension \(H_2\) under simulated Dirichlet RDPG graph distributions for varying separation parameter \(r\). Results are shown for LSF, ELSF, LpF, and~SMF using degree and betweenness centrality as filter functions, together with their concatenated vector representations based on GPL, PL, and~BF.}
    \label{fig:RDPG_power_H2}
\end{figure}
\unskip

\section{Real-World Graph~Dataset}
\label{app:datasets}

This section summarizes the real-world graph datasets used in Section~\ref{exp3}. Table~\ref{tab:real} presents the number of graphs, number of classes, and~average numbers of nodes and edges for each~dataset.

\begin{table}[H]
\centering
\caption{Summary of the real-world graph datasets used in
Section~\ref{exp3}, including the number of graphs,
number of classes, and average numbers of nodes and edges.}
\label{tab:real}

\resizebox{\textwidth}{!}{%
\begin{tabular}{lcccc}
\toprule
\textbf{Dataset}
& \textbf{No.\ of Graphs}
& \textbf{No.\ of Classes}
& \textbf{Avg.\ Nodes}
& \textbf{Avg.\ Edges} \\
\midrule
DHFR            & 756  & 2 & 42.43  & 44.54  \\
PROTEINS        & 1113 & 2 & 39.31  & 72.82  \\
IMDB-BINARY     & 1000 & 2 & 19.77  & 96.53  \\
REDDIT-BINARY   & 2000 & 2 & 429.60 & 497.80 \\
DD              & 1178 & 2 & 284.32 & 715.66 \\
COX2            & 467  & 2 & 41.22  & 43.45  \\
BZR             & 405  & 2 & 35.75  & 38.36  \\
IMDB-MULTI      & 1500 & 3 & 13.00  & 65.94  \\
ENZYMES         & 600  & 6 & 32.63  & 62.14  \\
REDDIT-MULTI-5K & 4999 & 5 & 508.52 & 594.87 \\
\bottomrule
\end{tabular}%
}
\end{table}


\section{Additional Hypothesis Testing Results for Real-World Graph~Datasets}
\label{app:h1_results}

This section presents the hypothesis testing results for homological dimension \(H_1\), complementing the \(H_0\) results reported in Section~\ref{exp3}. Table~\ref{tab:results_estat_grouped1} provides the corresponding results across the proposed filtration methods, filter functions, and~topological summaries, allowing a comparison of their performance across both homological~dimensions.

\begin{table}[H]
\centering
\caption{Summary of hypothesis testing results based on the energy
statistic for the real-world graph datasets in homological dimension
$H_1$. Results are reported for the three topological summary
functions (GPL, PL, and BF) across the four filtration methods
(LSF, ELSF, LpF, and SMF). Filtrations are induced using
betweenness and degree centrality as filter functions.
The columns $B_1$ and $D_1$ denote the $H_1$ topological
summaries obtained using betweenness and degree centrality,
respectively, while $C_1$ denotes their concatenated
representation. Test outcomes are denoted as follows:
\cmark\ indicates rejection of the null hypothesis ($p<0.05$),
\textcolor{blue}{*} indicates borderline evidence
($0.05 \leq p < 0.10$), and \xmark\ indicates failure to reject
the null hypothesis ($p \geq 0.10$).}
\label{tab:results_estat_grouped1}

\resizebox{\textwidth}{!}{%
\begin{tabular}{l*{12}{c}}
\toprule
\multirow{2}{*}{\textbf{Dataset}}
& \multicolumn{3}{c}{\textbf{LSF}}
& \multicolumn{3}{c}{\textbf{ELSF}}
& \multicolumn{3}{c}{\textbf{LpF}}
& \multicolumn{3}{c}{\textbf{SMF}} \\
\cmidrule(lr){2-4}
\cmidrule(lr){5-7}
\cmidrule(lr){8-10}
\cmidrule(lr){11-13}
& $\boldsymbol{B_1}$ & $\boldsymbol{D_1}$ & $\boldsymbol{C_1}$
& $\boldsymbol{B_1}$ & $\boldsymbol{D_1}$ & $\boldsymbol{C_1}$
& $\boldsymbol{B_1}$ & $\boldsymbol{D_1}$ & $\boldsymbol{C_1}$
& $\boldsymbol{B_1}$ & $\boldsymbol{D_1}$ & $\boldsymbol{C_1}$ \\
\midrule

\multicolumn{13}{c}{\textbf{GPL-$H_1$}} \\
\midrule
DHFR
& \cmark & \xmark & \cmark
& \cmark & \xmark & \cmark
& \cmark & \cmark & \cmark
& \cmark & \xmark & \cmark \\

PROTEINS
& \cmark & \cmark & \cmark
& \cmark & \cmark & \cmark
& \cmark & \cmark & \cmark
& \cmark & \cmark & \cmark \\

IMDB-BINARY
& \xmark & \cmark & \cmark
& \cmark & \cmark & \cmark
& \cmark & \cmark & \cmark
& \cmark & \cmark & \cmark \\

REDDIT-BINARY
& \cmark & \cmark & \cmark
& \cmark & \cmark & \cmark
& \cmark & \cmark & \cmark
& \cmark & \cmark & \cmark \\

DD
& \cmark & \cmark & \cmark
& \cmark & \cmark & \cmark
& \cmark & \cmark & \cmark
& \cmark & \cmark & \cmark \\

COX2
& \cmark & \cmark & \cmark
& \cmark & \cmark & \cmark
& \cmark & \cmark & \cmark
& \cmark & \cmark & \cmark \\

BZR
& \cmark & \cmark & \cmark
& \cmark & \cmark & \cmark
& \cmark & \cmark & \cmark
& \cmark & \cmark & \cmark \\

ENZYMES
& \cmark & \cmark & \cmark
& \cmark & \cmark & \cmark
& \cmark & \cmark & \cmark
& \cmark & \cmark & \cmark \\

IMDB-MULTI
& \cmark & \xmark & \xmark
& \textcolor{blue}{*} & \cmark & \cmark
& \cmark & \cmark & \cmark
& \xmark & \cmark & \cmark \\

REDDIT-MULTI-5K
& \cmark & \cmark & \cmark
& \cmark & \cmark & \cmark
& \cmark & \cmark & \cmark
& \cmark & \cmark & \cmark \\
\midrule

\multicolumn{13}{c}{\textbf{PL-$H_1$}} \\
\midrule
DHFR
& \cmark & \cmark & \cmark
& \cmark & \cmark & \cmark
& \cmark & \cmark & \cmark
& \cmark & \cmark & \cmark \\

PROTEINS
& \xmark & \cmark & \cmark
& \xmark & \cmark & \cmark
& \cmark & \cmark & \cmark
& \cmark & \cmark & \cmark \\

IMDB-BINARY
& \cmark & \cmark & \cmark
& \cmark & \cmark & \cmark
& \cmark & \cmark & \cmark
& \cmark & \cmark & \cmark \\

REDDIT-BINARY
& \cmark & \cmark & \cmark
& \cmark & \cmark & \cmark
& \cmark & \cmark & \cmark
& \xmark & \cmark & \cmark \\

DD
& \xmark & \cmark & \xmark
& \xmark & \cmark & \xmark
& \xmark & \cmark & \xmark
& \cmark & \cmark & \cmark \\

COX2
& \cmark & \cmark & \cmark
& \cmark & \cmark & \cmark
& \cmark & \cmark & \cmark
& \cmark & \cmark & \cmark \\

BZR
& \xmark & \cmark & \cmark
& \xmark & \cmark & \cmark
& \xmark & \cmark & \cmark
& \xmark & \cmark & \cmark \\

ENZYMES
& \cmark & \cmark & \cmark
& \cmark & \cmark & \cmark
& \cmark & \cmark & \cmark
& \cmark & \cmark & \cmark \\

IMDB-MULTI
& \xmark & \xmark & \xmark
& \xmark & \xmark & \xmark
& \xmark & \cmark & \cmark
& \xmark & \cmark & \xmark \\

REDDIT-MULTI-5K
& \cmark & \cmark & \cmark
& \cmark & \cmark & \cmark
& \cmark & \cmark & \cmark
& \xmark & \cmark & \cmark \\
\midrule

\multicolumn{13}{c}{\textbf{BF-$H_1$}} \\
\midrule
DHFR
& \cmark & \cmark & \cmark
& \cmark & \cmark & \cmark
& \cmark & \cmark & \cmark
& \cmark & \cmark & \cmark \\

PROTEINS
& \cmark & \cmark & \cmark
& \cmark & \cmark & \cmark
& \cmark & \cmark & \cmark
& \cmark & \cmark & \cmark \\

IMDB-BINARY
& \cmark & \cmark & \cmark
& \cmark & \cmark & \cmark
& \cmark & \cmark & \cmark
& \cmark & \cmark & \cmark \\

REDDIT-BINARY
& \cmark & \cmark & \cmark
& \cmark & \cmark & \cmark
& \cmark & \cmark & \cmark
& \cmark & \cmark & \cmark \\

DD
& \cmark & \cmark & \cmark
& \cmark & \cmark & \cmark
& \cmark & \cmark & \cmark
& \cmark & \cmark & \cmark \\

COX2
& \cmark & \cmark & \cmark
& \cmark & \cmark & \cmark
& \cmark & \cmark & \cmark
& \cmark & \cmark & \cmark \\

BZR
& \textcolor{blue}{*} & \cmark & \cmark
& \textcolor{blue}{*} & \cmark & \cmark
& \xmark & \cmark & \cmark
& \textcolor{blue}{*} & \cmark & \cmark \\

ENZYMES
& \cmark & \cmark & \cmark
& \cmark & \cmark & \cmark
& \cmark & \cmark & \cmark
& \cmark & \cmark & \cmark \\

IMDB-MULTI
& \xmark & \cmark & \textcolor{blue}{*}
& \cmark & \cmark & \cmark
& \cmark & \cmark & \cmark
& \cmark & \cmark & \cmark \\

REDDIT-MULTI-5K
& \cmark & \cmark & \cmark
& \cmark & \cmark & \cmark
& \cmark & \cmark & \cmark
& \xmark & \cmark & \cmark \\
\bottomrule

\end{tabular}%
}
\end{table}

\section{Additional Chest X-Ray~Results}
\label{app:image_resolution}

\begin{figure}[H]
    
    \includegraphics[width=1.0\linewidth]{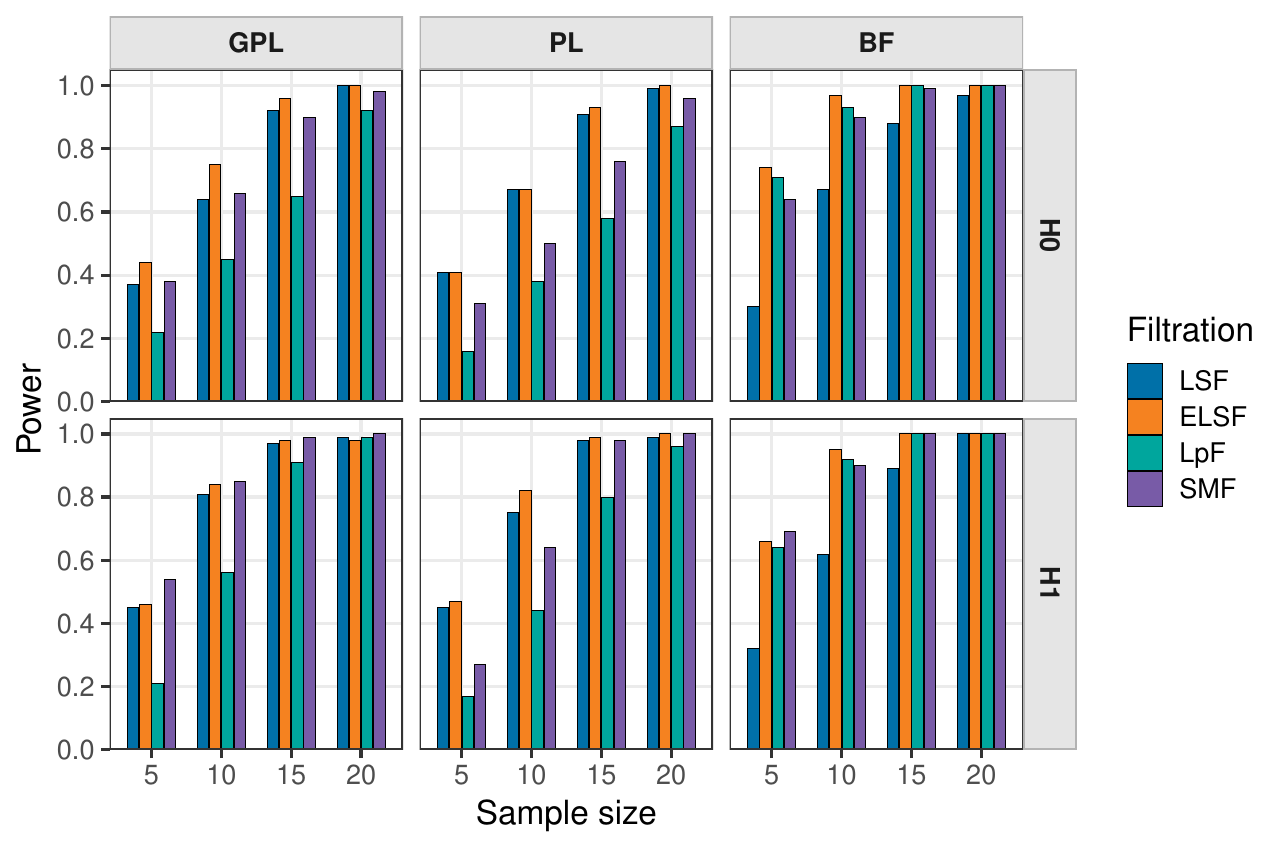}
    \caption{
Estimated power of the energy-based permutation test for chest X-ray image data as sample size increases using images with resolution $64 \times 64$ pixels at significance level $\alpha = 0.05$. Results are shown for four filtration methods (LSF, ELSF, LpF, and~SMF) and three topological summaries (GPL, PL, and~BF) across homological dimensions $H_0$ (top row) and $H_1$ (bottom row).
}
    \label{fig11}
\end{figure}

\begin{figure}[H]
    
    \includegraphics[width=1.0\linewidth]{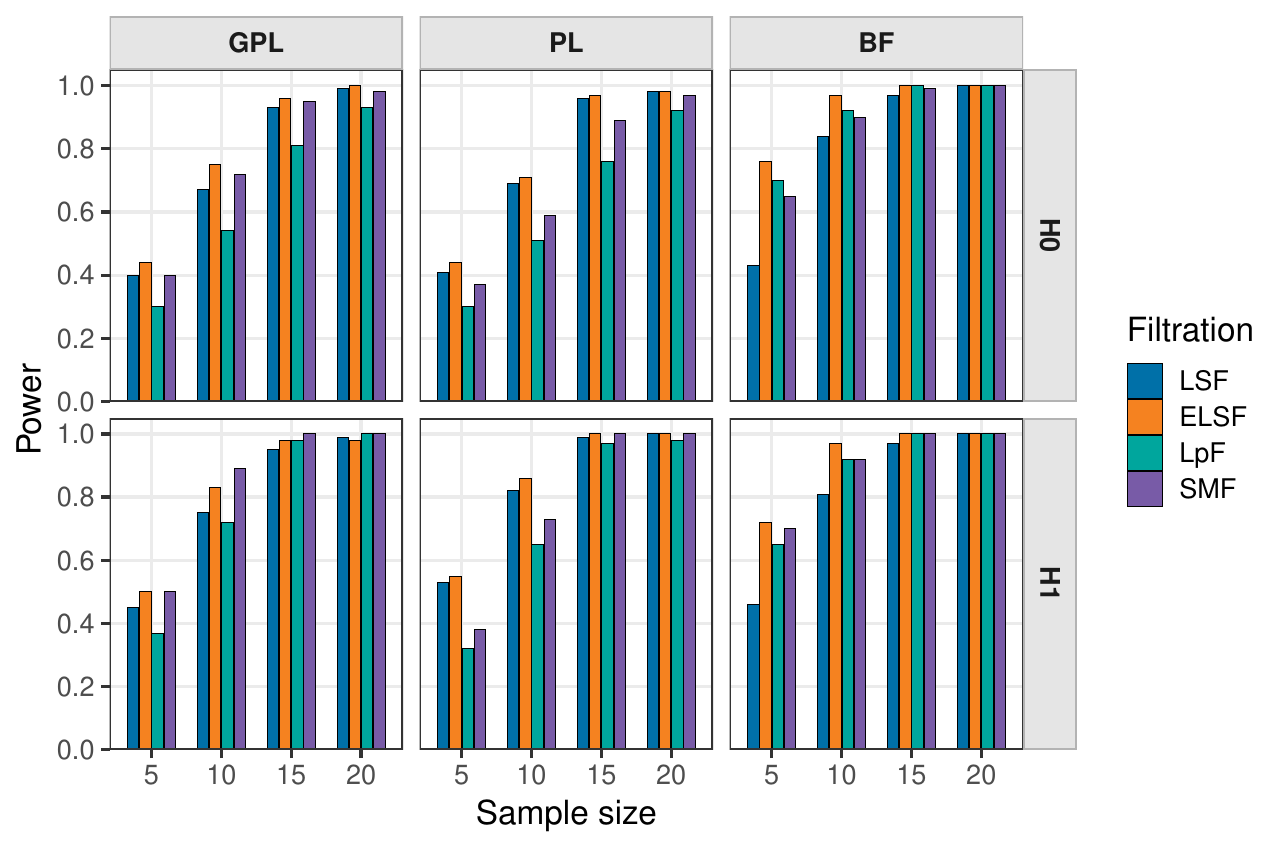}
    \caption{
Estimated power of the energy-based permutation test for chest X-ray image data as sample size increases using images with resolution $128 \times 128$ pixels at significance level $\alpha = 0.05$. Results are shown for four filtration methods (LSF, ELSF, LpF, and~SMF) and three topological summaries (GPL, PL, and~BF) across homological dimensions $H_0$ (top row) and $H_1$ (bottom row).
}
    \label{fig12}
\end{figure}

The experiments presented in Section~\ref{x-ray} were conducted using chest X-ray images resized to \(256 \times 256\) pixels. To~investigate the effect of image resolution on the proposed framework, we additionally repeated the experiments using images resized to \(64 \times 64\) and \(128 \times 128\) pixels. Figures~\ref{fig11} and~\ref{fig12} present the corresponding empirical power curves for these additional image resolutions, demonstrating the performance of the proposed filtration methods under lower-resolution~settings.

\unskip

\bibliographystyle{unsrtnat}
\bibliography{graphtime}

\end{document}